\documentclass[sigconf]{acmart}
\AtBeginDocument{%
  }
\usepackage{bm}
\usepackage{csquotes}
\usepackage{multirow}
\usepackage[ruled, algo2e]{algorithm2e}
\usepackage[capitalize,noabbrev]{cleveref}
\copyrightyear{2026}
\acmYear{2026}
\setcopyright{cc}
\setcctype{by-nc-nd}
\acmConference[KDD '26]{Proceedings of the 32nd ACM SIGKDD Conference on Knowledge Discovery and Data Mining V.2}{August 09--13, 2026}{Jeju Island, Republic of Korea}
\acmBooktitle{Proceedings of the 32nd ACM SIGKDD Conference on Knowledge Discovery and Data Mining V.2 (KDD '26), August 09--13, 2026, Jeju Island, Republic of Korea}
\acmDOI{10.1145/3770855.3818009}
\acmISBN{979-8-4007-2259-2/2026/08}

\acmSubmissionID{v2rtp2474}

\begin{document}

\title{Towards Stable, Globally Expressive Graph Representations with Laplacian Eigenvectors}


\author{Junru Zhou}
\email{zml72062@126.com}
\affiliation{
  \institution{Institute for Artificial Intelligence, Peking University}
  \city{Beijing}
  \country{China}
}

\author{Cai Zhou}
\email{caiz428@mit.edu}
\affiliation{
\institution{Department of EECS, Massachusetts Institute of Technology}
\city{Cambridge}
\state{Massachusetts}
\country{USA}
}

\author{Xiyuan Wang}
\email{wangxiyuan@pku.edu.cn}
\affiliation{
  \institution{Institute for Artificial Intelligence, Peking University}
  \city{Beijing}
  \country{China}
}

\author{Pan Li}
\email{panli@gatech.edu}
\affiliation{
\institution{School of ECE, Georgia Institute of Technology}
\city{Atlanta}
\state{Georgia}
\country{USA}
}

\author{Muhan Zhang}
\email{muhan@pku.edu.cn}
\affiliation{
  \institution{Institute for Artificial Intelligence, Peking University}
  \city{Beijing}
  \country{China}
}

\renewcommand{\shortauthors}{Zhou et al.}

\begin{abstract}
A popular way to improve the expressive power of graph neural networks (GNNs) is to use Laplacian eigenvectors as additional node features, since they can serve both as structural identifiers and global coordinates of nodes. Properly handling the orthogonal group symmetry among eigenvectors is crucial for the stability and generalizability of Laplacian eigenvector augmented GNNs. Previous studies have shown that using a naive $O(p)$-group invariant encoder for each $p$-dimensional eigenspace often leads to expressivity loss and numerical instability. In this paper, we propose a novel method exploiting Laplacian eigenvectors to generate \emph{stable} and globally \emph{expressive} graph representations. The main difference from previous works is that (i) our method utilizes \textbf{learnable} $O(p)$-invariant representations for each Laplacian eigenspace of dimension $p$, which are built upon powerful orthogonal group equivariant neural network layers already well studied in the literature, and that (ii) our method deals with numerically close eigenvalues in a \textbf{smooth} fashion, ensuring its better robustness against perturbations. Experiments on various graph learning benchmarks witness the competitive performance of our method, especially its great potential to learn global properties of graphs.
\end{abstract}

\received{20 February 2007}
\received[revised]{12 March 2009}
\received[accepted]{5 June 2009}

\maketitle

\section{Introduction}\label{sect:intro}


Graph neural networks (GNNs) have proved successful on a plethora of learning tasks over graphs~\citep{wu2020comprehensive, zhou2020graph}, spanning across domains such as chemistry~\citep{deshpande2002automated, jin2018junction, reiser2022graph}, biology~\citep{stokes2020deep, zitnik2017predicting, zitnik2018modeling}, social recommendations~\citep{ying2018graph} or electronic design automation~\citep{lopera2021survey}. One of the most popularly adopted GNN architecture is message passing neural network (MPNN), which maintains a representation vector $\bm{h}_u$ for each node $u$, and iteratively updates it by gathering information from the neighboring nodes of $u$. Despite its relative simplicity and efficiency, several weaknesses severely limit its performance. One important problem is its \textbf{limited expressive power}, referring to the fact that MPNNs often fail to distinguish between two non-isomorphic graphs, or two structurally different nodes with similar neighborhood configuration~\citep{xu2018powerful, zhang2021labeling}. Another issue is its \textbf{inability to capture global properties} of graphs, meaning that it cannot truthfully learn long-range interactions within a graph, due to ``oversquashing'' that occurs as a result of multiple message passing steps~\citep{alon2020bottleneck, dwivedi2022long}. In this paper, we refer to the former problem as a lack of \textbf{local} expressive power, while the latter as one concerning \textbf{global} expressive power.

A great number of works have attempted to tackle the aforementioned weaknesses of MPNNs, which we review in Section~\ref{sect:rel_work}. For now, we restrict our discussion to one specific approach---using Laplacian eigenvectors as node feature augmentations. We are particularly interested in such an approach since Laplacian eigenvectors provide a solution to alleviate \textbf{both} issues we raise above. First, Laplacian eigenvectors provably contain rich \textbf{local structural information}~\citep{cvetkovic1997eigenspaces, FURER20102373, Rattan_2023}, and can thus serve as additional node labels, making it easier for a GNN to separate nodes that are otherwise similar. Furthermore, they can reflect the absolute position of each node within the graph~\citep{von2007tutorial}, making GNNs aware of potential \textbf{long-range interactions}. Indeed, a vast literature has regarded Laplacian eigenvectors as \emph{graph positional encodings}, which play important roles both in MPNNs and graph transformers~\citep{dwivedi2021graph, dwivedi2023benchmarking, rampavsek2022recipe, ying2021transformers}.

Although Laplacian eigenvectors provide a promising approach to expressive graph representation learning, there are some well-known constraints that must be taken into account for their reliable use. The first is \textbf{orthogonal-group invariance}. As is first pointed out by~\citet{wang2022equivariant,lim2022sign}, given a Laplacian $\bm{L}$, its eigen-decomposition is in general not unique. In fact, assuming that $\bm{v}_1, \ldots, \bm{v}_p$ are $p$ mutually orthogonal normalized eigenvectors of a Laplacian $\bm{L}$ that correspond to the same eigenvalue $\lambda$, then so are $\bm{v}'_1, \ldots, \bm{v}'_p$, as long as the two groups of eigenvectors can be associated via a $p\times p$ orthogonal matrix $\bm{Q}$, namely $\bm{V}'=\bm{V}\cdot \bm{Q}$ where $\bm{V}=(\bm{v}_1,\ldots, \bm{v}_p)$ and $\bm{V}'=(\bm{v}'_1,\ldots, \bm{v}'_p)$. One must ensure that the network output is invariant to such orthogonal transformations, so as to produce identical representations for identical (i.e., isomorphic) graphs. Another related but stricter constraint is \textbf{stability}, which, as formally defined by~\citet{huang2023stability}, demands that network outputs should be close when the input graph undergoes small perturbations. It is easy to see that orthogonal-group invariance is a special case of stability in which the strength of perturbation approaches zero.

To ensure orthogonal-group invariance (and furthermore, stability),~\citet{lim2022sign} and~\citet{huang2023stability} both propose to extract spectral information from \emph{inner products} between Laplacian eigenvectors---namely, $\bm{VV}^T$ with $\bm{V}=(\bm{v}_1,\ldots, \bm{v}_p)$ being the matrix consisting of $p$ mutually orthogonal eigenvectors within an eigenspace of dimension $p$---instead of the eigenvectors $(\bm{v}_1,\ldots, \bm{v}_p)$ themselves. Despite being provably invariant to $O(p)$ transformations and even stable (with carefully designed network architectures), such learning architectures based on inner products are not flexible enough, and may lose much of the rich structural and positional information carried by vanilla Laplacian eigenvectors. Earlier than the above two works,~\citet{wang2022equivariant} has proposed a special message passing operation in which only the norms of differences between rows of $\bm{V}$ are used, and proved its stability. However, this method even dismisses important eigenvalue information by treating Laplacian eigenvectors from different eigenspaces uniformly. Given the limitations of existing methods to utilize Laplacian eigenvectors, a natural question is \textbf{whether we can recover the information inherent in vanilla Laplacian eigenvectors while ensuring stability}. To make the question even more general, we may ask:

\begin{displayquote}
\emph{What is the representational limit of graph learning methods exploiting Laplacian eigenvectors, given the stability constraint?}
\end{displayquote}

In this paper, we attempt to partially answer the general question posed above. Our main contributions are summarized below. 

\begin{itemize}
    \item We propose vanilla orthogonal group equivariant augmentation (\textbf{Vanilla OGE-Aug}), a novel method exploiting Laplacian eigenvectors to produce node feature augmentations. We make use of powerful \textbf{invariant point cloud networks} (for example, Tensor Field Network~\citep{thomas2018tensor} and its variant~\citep{finkelshtein2022simple}) to process the Laplacian eigenvectors, which enables construction of node feature augmentations much more expressive than previous ones based on inner products between eigenvectors.

    \item We theoretically prove that Vanilla OGE-Aug, combined with an MPNN, can lead to \textbf{universal representations} of graphs, as long as the invariant point cloud networks we use are expressive enough. Previous works have theoretically guaranteed the expressive power of several specific invariant point cloud networks, which lays the foundations for the practicality of our theoretical result.
    
    \item Although Vanilla OGE-Aug can be maximally expressive, it unfortunately lacks stability. We then propose a \textbf{smooth} variant of Vanilla OGE-Aug, namely \textbf{OGE-Aug}, by trading expressive power for better stability. Our approach is to use a series of ``soft'' masks to filter Laplacian eigenvectors that belong to different eigenspaces, instead of hard-splitting and separately processing them. We theoretically prove the stability of OGE-Aug, and evaluate its empirical performance on various real-world graph datasets. The results indicate that our method not only shows competitive performance on popular graph benchmarks, but is surprisingly good at learning \textbf{global properties of graphs}.
\end{itemize}

\section{Preliminaries}\label{sect:prelim}

We use $\mathcal{G}$ to denote the set of all simple, undirected graphs. For a graph $G\in\mathcal{G}$, its node set and edge set are denoted by $\mathcal{V}(G)$ and $\mathcal{E}(G)$ respectively. If $G$ has $n$ nodes labeled by $1,\ldots, n$ respectively, its adjacency matrix is defined as $\bm{A}_G\in\{0,1\}^{n\times n}$ in which $A_{Gij}=1$ if and only if $\{i,j\}$ is an edge. Further, if $G$ has node features, the node features are represented by a matrix $\bm{X}_G\in\mathbb{R}^{n\times d}$ whose $i$-th row corresponds to the feature of node $i$. 

The Laplacian of graph $G$ is defined as $\bm{L}_G=\bm{D}_G-\bm{A}_G$, in which $\bm{D}_G=\mathrm{diag}(d_G(1),\ldots, d_G(n))$, with $d_G(i)$ being the degree of node $i$ ($i=1,\ldots, n$). It is not hard to see that for any $G\in\mathcal{G}$, $\bm{L}_G$ is positive semi-definite. Therefore, all eigenvalues of $\bm{L}_G$ are real non-negative. One may also verify that 0 is always an eigenvalue of $\bm{L}_G$ (thus being the smallest eigenvalue of $\bm{L}_G$). If $\bm{L}_G$ has an eigenvalue $\lambda$ with multiplicity $\mu$, the linear subspace spanned by the $\mu$ mutually orthogonal eigenvectors of $\bm{L}_G$ corresponding to $\lambda$ is called an eigenspace of $\bm{L}_G$ with dimension $\mu$.

Let $\bm{A}\in\mathbb{R}^{n\times n}$. $\bm{A}$ is said to be orthogonal if $\bm{AA}^T=\bm{A}^T\bm{A}=\bm{I}$, with $\bm{I}$ being the identity matrix. Given a positive integer $n$, we use $O(n)$ to denote the set of all orthogonal matrices of shape $n\times n$. A 0-1 matrix $\bm{A}\in\{0,1\}^{n\times n}$ is said to be a permutation matrix if each of its rows and columns has exactly one $1$-element. Let $S_n$ be the set of all permutation matrices of shape $n\times n$. It's easy to see that $S_n\subseteq O(n)$.

To simplify our discussion below, we further introduce the following shorthands:

\begin{itemize}
    \item Assume $\{\bm{V}_1,\ldots, \bm{V}_k\}\subset\mathbb{R}^{n\times p}$ is a set of $n\times p$ matrices, in which $\bm{V}_j$ can be row-wise decomposed as $\bm{V}_j=(\bm{v}_{j1},\ldots, \bm{v}_{jn})^T$, each $\bm{v}_{ji}\in\mathbb{R}^p$, $i=1,\ldots, n$. Further let $g$ be a set function, namely $g:2^{\mathbb{R}^p}\rightarrow\mathbb{R}$. Then we use $g(\{\bm{V}_1,\ldots, \bm{V}_k\})\in\mathbb{R}^n$ to denote the vector whose $i$-th component equals $g(\{\bm{v}_{1i},\ldots, \bm{v}_{ki}\})$, for $i=1,\ldots, n$.
    \item Given $\bm{V}_1\in\mathbb{R}^{n\times p_1},\ldots, \bm{V}_k\in\mathbb{R}^{n\times p_k}$, let $\mathrm{concat}\left[\bm{V}_1,\ldots,\bm{V}_k\right]\in\mathbb{R}^{n\times (p_1+\cdots+p_k)}$ be the concatenation of $\bm{V}_1,\ldots, \bm{V}_k$ along the row dimension.
\end{itemize}

\section{Universal graph representation with Laplacian eigenvectors}\label{sect:univ}

Despite the great number of works showing the efficacy of using Laplacian eigenvectors in graph learning tasks, few~\citep{FURER20102373, Rattan_2023} have studied theoretically their expressiveness upper-bound---namely, \textbf{to what extent can the information of a graph be learned, merely from its Laplacian eigenvectors?} This is a weaker version of the general question we pose in Section~\ref{sect:intro}, with the stability constraint removed. In this section, we will show that the answer to this weaker question is rather optimistic: ignoring the stability constraint, Laplacian eigenvectors can actually lead to \textbf{universal representations} of graphs. To reach the point, we start by reconsidering the problem of finding universal graph representations from the perspective of Laplacian eigenvalues and eigenvectors (Proposition~\ref{prop:universal_equiv}), and then give a concrete construction of such universal representation (Proposition~\ref{prop:construction}).


We first present the definition for universal representations of graphs.
\begin{definition}[Universal representation]\label{def:universal}
    Let $f$ be a function mapping each pair $(G,\bm{X}_G)$ to a real value $f(G,\bm{X}_G)\in\mathbb{R}$, where $G\in\mathcal{G}$ is a graph and $\bm{X}_G\in \mathbb{R}^{|\mathcal{V}(G)|\times d}$ stands for node features accompanied with $G$. Further let $\bm{A}_G$ be the adjacency matrix of graph $G$. The function $f$ is said to be a \textbf{universal representation} if the following condition holds: for any two pairs $(G, \bm{X}_G)$ and $(H, \bm{X}_H)$, $f(G,\bm{X}_G)=f(H,\bm{X}_H)$ if and only if $\exists \bm{P}\in S_{|\mathcal{V}(G)|}$, $\bm{A}_G=\bm{PA}_H\bm{P}^T, \bm{X}_G = \bm{PX}_H$.
\end{definition}
In other words, $f$ should produce equal outputs only for graphs that are identical up to a permutation of nodes.

Next, we will associate the concept of universal representations with eigendecompositions of graph Laplacians. Let $\bm{L}_G$ be the Laplacian of $G\in\mathcal{G}$. Due to the properties of graph Laplacians (stated in Section~\ref{sect:prelim}), we may assume that $\bm{L}_G$ has $K$ distinct real eigenvalues $\lambda_1,\ldots, \lambda_K$, with $0=\lambda_1<\lambda_2<\cdots<\lambda_K$. We further use $\mu_j$ to denote the multiplicity of eigenvalue $\lambda_j$, and $\bm{V}_j\in\mathbb{R}^{|\mathcal{V}(G)|\times \mu_j}$ the set of mutually orthogonal normalized eigenvectors corresponding to $\lambda_j$ (each column of $\bm{V}_j$ being an eigenvector that has $L^2$-norm scaled to 1), for $j=1,\ldots, K$. Following~\citet{FURER20102373}, the \emph{spectrum} of $G$ is defined as 
\begin{gather}
    \mathrm{Spec}\ G = ((\lambda_1, \mu_1), (\lambda_2, \mu_2),\ldots, (\lambda_K, \mu_K)).
\end{gather}

Given the above notations, the following proposition is straightforward.
\begin{proposition}\label{prop:universal_equiv}
    Let $G,H\in\mathcal{G}$ with $|\mathcal{V}(G)|=|\mathcal{V}(H)|$. Let $\bm{A}_G$ and $\bm{A}_H$ be their adjacency matrices respectively. The following two statements are equivalent:\\
    (i) $\exists \bm{P}\in S_{|\mathcal{V}(G)|}, \bm{A}_G=\bm{PA}_H\bm{P}^T$.\\
    (ii) Both of the following conditions hold.
    \begin{itemize}
        \item $\mathrm{Spec}\ G=\mathrm{Spec}\ H$.
        \item Let the spectrum of $G$ (and thus $H$) be $((\lambda_1, \mu_1), \ldots, (\lambda_K, \mu_K))$, and $\bm{V}_j, \bm{V}'_j\in\mathbb{R}^{|\mathcal{V}(G)|\times \mu_j}$ be sets of mutually orthogonal normalized eigenvectors belonging to $G,H$ respectively, both corresponding to eigenvalue $\lambda_j$, for $j=1,\ldots, K$. There exists $\bm{P}\in S_{|\mathcal{V}(G)|}$ and $\bm{Q}_j\in O(\mu_j)$ ($j=1,\ldots, K$), such that $\bm{V}_j=\bm{PV}'_j\bm{Q}_j$.
    \end{itemize}
\end{proposition}

\begin{proof}
    We denote $n=|\mathcal{V}(G)|$. Let $\bm{L}_G$ and $\bm{L}_H$ be the Laplacians of $G$ and $H$, respectively. We first show that statement (i) is equivalent to the following: $\exists \bm{P}\in S_n, \bm{L}_G=\bm{PL}_H\bm{P}^T$. By definition of permutation matrices, for any $\bm{P}\in S_n$ there exists a bijective function $p:\{1,\ldots, n\}\rightarrow\{1,\ldots, n\}$ such that $P_{ij}=1_{p(i)=j}$. Therefore, we have $\bm{L}_G=\bm{PL}_H\bm{P}^T\Leftrightarrow L_{Gij}=L_{Hp(i)p(j)}$. Since the off-diagonal part of $\bm{L}_G$ (or $\bm{L}_H$) is $-\bm{A}_G$ (or $-\bm{A}_H$), $L_{Gij}=L_{Hp(i)p(j)}$ implies $A_{Gij}=A_{Hp(i)p(j)}$. Thus $\bm{A}_G=\bm{P}\bm{A}_H\bm{P}^T$ follows from $\bm{L}_G=\bm{PL}_H\bm{P}^T$. To see the other direction, notice that given $\bm{A}_G=\bm{PA}_H\bm{P}^T$ or $A_{Gij}=A_{Hp(i)p(j)}$, we have
    \begin{align}
        \notag (\bm{PD}_H\bm{P}^T)_{ij} &= D_{Hp(i)p(i)}1_{i=j}=\sum_{k=1}^n A_{Hp(i)p(k)}1_{i=j}\\&=\sum_{k=1}^n A_{Gik}1_{i=j} = D_{Gij},
    \end{align}
    or simply $\bm{D}_G=\bm{PD}_H\bm{P}^T$. Thus $\bm{L}_G=\bm{PL}_H\bm{P}^T$.

    Next, we prove that statement (ii) is equivalent to $\exists \bm{P}\in S_n, \bm{L}_G=\bm{PL}_H\bm{P}^T$. Assuming that statement (ii) is true, one may make use of the identities
    \begin{gather}
        \bm{L}_G = \sum_{j=1}^K \lambda_j\bm{V}_j\bm{V}_j^T,\quad \bm{L}_H = \sum_{j=1}^K \lambda_j\bm{V}'_j\bm{V}'^T_j,
    \end{gather}
    to observe that $\bm{L}_G=\bm{PL}_H\bm{P}^T$. To see the other direction, notice that $\bm{L}_G=\bm{PL}_H\bm{P}^T$ implies that $\bm{L}_G$ and $\bm{L}_H$ are similar, and thus $\mathrm{Spec}\ G=\mathrm{Spec}\ H$ as similar matrices share the set of eigenvalues combined with their corresponding multiplicities. Moreover, if the columns of $\bm{V}'_j$ constitute the set of mutually orthogonal normalized eigenvectors of $\bm{L}_H$ corresponding to eigenvalue $\lambda_j$, then the columns of $\bm{PV}'_j$ contain mutually orthogonal normalized eigenvectors of $\bm{L}_G$ corresponding to the same eigenvalue, for $j=1,\ldots, K$. Therefore, each column of $\bm{V}_j$ must be a linear combination of columns of $\bm{PV}'_j$, namely
    \begin{gather}
        \bm{V}_j = \bm{PV}'_j\bm{Q}_j,
    \end{gather}
    for some $\bm{Q}_j\in\mathbb{R}^{\mu_j\times \mu_j}$. Further imposing the constraint that $\bm{V}_j^T\bm{V}_j=\bm{I}_{\mu_j\times \mu_j}$ yields $\bm{Q}_j^T\bm{Q}_j = \bm{I}_{\mu_j\times \mu_j}$, or $\bm{Q}_j\in O(\mu_j)$. 
\end{proof}

Proposition~\ref{prop:universal_equiv} implies that in order to find universal representations of a graph, it may be helpful to find a sufficiently expressive representation for each of its Laplacian eigenspace. Nevertheless, such representation must stay invariant under actions of $O(p)$-group elements for an eigenspace of dimension $p$, due to the existence of arbitrary $\bm{Q}_j$ matrices ($j=1,\ldots, K$). Thus, we are motivated to define as following an $O(p)$-invariant universal representation.

\begin{definition}[$O(p)$-invariant universal representation]\label{def:op-inv}
    Let $f:\bigcup_{n=0}^\infty \mathbb{R}^{n\times p}\rightarrow \bigcup_{n=0}^\infty \mathbb{R}^{n\times 1}$. Given an input $\bm{V}\in\mathbb{R}^{n\times p}$, $f$ outputs a column vector $f(\bm{V})\in\mathbb{R}^{n\times 1}$. The function $f$ is said to be an \textbf{$\bm{O(p)}$-invariant universal representation} if given $\bm{V}, \bm{V}'\in\mathbb{R}^{n\times p}$ and $\bm{P}\in S_n$, the following two conditions are equivalent: (i) $f(\bm{V}) = \bm{P}f(\bm{V}')$; (ii) $\exists \bm{Q}\in O(p)$, such that $\bm{V}=\bm{PV}'\bm{Q}$.
\end{definition}

By Definition~\ref{def:op-inv}, an $O(p)$-invariant universal representation injectively assigns a point-wise value to a point set embedded in $\mathbb{R}^p$, while ensuring invariance to global $O(p)$ rotations and equivariance to point permutations. Such networks have been named \emph{universal point cloud networks}, whose design has been intensively studied, as we will survey in Section~\ref{sect:rel_work}. 

We still need another definition which follows~\citet{zaheer2017deep}.

\begin{definition}[Universal set representation]\label{def:set-inv}
    Let $\mathcal{X}$ be a non-empty set. A function $f:2^{\mathcal{X}}\rightarrow\mathbb{R}$ is said to be a \textbf{universal set representation} if $\forall X_1, X_2\in 2^{\mathcal{X}}$, $f(X_1)=f(X_2)$ if and only if the two sets $X_1$ and $X_2$ are equal.
\end{definition}

We remark that the problem of finding a universal set representation, at least for finite subsets of a countable universe $\mathcal{X}$, has been fully addressed by~\citet{zaheer2017deep}, using the deep set architecture they propose.

With Definitions~\ref{def:op-inv}~and~\ref{def:set-inv}, we are now ready to present our main result on constructing universally expressive graph representations. 

\begin{proposition}\label{prop:construction}
    For each $p=1,2,\ldots$, let $f_p$ be an $O(p)$-invariant universal representation function. Further let $g:2^{\mathbb{R}^3}\rightarrow\mathbb{R}$ be a universal set representation. Then the following function
    \begin{small}
    \begin{align}
        \notag r(G, \bm{X}_G) &= \mathrm{GNN}\Big(\bm{A}_G, \mathrm{concat}\Big[\bm{X}_G, g\\
        &\left(\left\{
        \mathrm{concat}\left[\mu_j\mathbf{1}_n, \lambda_j\mathbf{1}_n, f_{\mu_j}(\bm{V}_j)\right]\right\}_{j=1}^K\right)\Big]\Big)\label{eq:construction}
    \end{align}
    \end{small}
    is a universal representation (by Definition~\ref{def:universal}). Here $n=|\mathcal{V}(G)|$, $((\lambda_1,\mu_1),\ldots, (\lambda_K,\mu_K))$ is the spectrum of $G$, and $\bm{V}_j\in\mathbb{R}^{n\times \mu_j}$ are the $\mu_j$ mutually orthogonal normalized eigenvectors of $\bm{L}_G$ corresponding to $\lambda_j$. We denote $\mathbf{1}_n$ an all-1 vector of shape $n\times 1$. $\mathrm{GNN}$ is a maximally expressive MPNN such as the one proposed in~\citep{xu2018powerful}. 
\end{proposition}

\begin{proof}
    By Definition~\ref{def:universal}, we only need to prove that $r(G,\bm{X}_G)=r(H,\bm{X}_H)$ if and only if $\exists \bm{P}\in S_n$ such that $\bm{A}_G=\bm{PA}_H\bm{P}^T$ and $\bm{X}_G=\bm{PX}_H$, for any two graphs $G,H$ with accompanying node features $\bm{X}_G,\bm{X}_H$. By Proposition~\ref{prop:universal_equiv}, the latter condition is equivalent to the conjunction of the following:
    \begin{enumerate}
        \item $\mathrm{Spec}\ G = \mathrm{Spec}\ H$.
        \item $\exists \bm{P}\in S_n$ and $\bm{Q}_j\in O(\mu_j)$ ($j=1,\ldots, K$), such that $\bm{X}_G=\bm{PX}_H$, and $\bm{V}_j = \bm{PV}'_j\bm{Q}_j$, for $j=1,\ldots, K$.
    \end{enumerate}
    Our notations follow those in Proposition~\ref{prop:universal_equiv}. Now, given that the above two conditions are true, we immediately get $f_{\mu_j}(\bm{V}_j)=\bm{P}f_{\mu_j}(\bm{V}'_j)$ due to the fact that $f_{\mu_j}$ is an $O(\mu_j)$-invariant universal representation. Thus, we have
    \begin{align}
        \mathrm{concat}\left[\mu_j\mathbf{1}_n, \lambda_j\mathbf{1}_n, f_{\mu_j}(\bm{V}_j)\right]=\bm{P}\ \mathrm{concat}\left[\mu_j\mathbf{1}_n, \lambda_j\mathbf{1}_n, f_{\mu_j}(\bm{V}'_j)\right].\label{eq:_concat_equal}
    \end{align}
    Similarly, since $g$ operates on individual rows of set elements, the permutation matrix $\bm{P}$ passes through the operation of $g$. Therefore,
    \begin{align}\label{eq:_g_set_equal}
    \notag &g\left(\left\{\mathrm{concat}\left[\mu_j\mathbf{1}_n, \lambda_j\mathbf{1}_n, f_{\mu_j}(\bm{V}_j)\right]\right\}_{j=1}^K\right) \\ = &\bm{P}g\left(\left\{\mathrm{concat}\left[\mu_j\mathbf{1}_n, \lambda_j\mathbf{1}_n, f_{\mu_j}(\bm{V}'_j)\right]\right\}_{j=1}^K\right).
    \end{align}
    If we let 
    \begin{gather}
        \bm{X}'_G = \mathrm{concat}\left[\bm{X}_G, g\left(\left\{\mathrm{concat}\left[\mu_j\mathbf{1}_n, \lambda_j\mathbf{1}_n, f_{\mu_j}(\bm{V}_j)\right]\right\}_{j=1}^K\right)\right],\label{eq:_xg_prime_def}\\
        \bm{X}'_H = \mathrm{concat}\left[\bm{X}_H, g\left(\left\{\mathrm{concat}\left[\mu_j\mathbf{1}_n, \lambda_j\mathbf{1}_n, f_{\mu_j}(\bm{V}'_j)\right]\right\}_{j=1}^K\right)\right],\label{eq:_xh_prime_def}
    \end{gather}
    then $\bm{X}'_G = \bm{PX}'_H$. Since message passing GNNs are invariant with respect to node permutations, we know that
    \begin{align}
        \notag r(G,\bm{X}_G)&=\mathrm{GNN}(\bm{A}_G, \bm{X}'_G) = \mathrm{GNN}(\bm{PA}_H\bm{P}^T, \bm{PX}'_H)\\ &= \mathrm{GNN}(\bm{A}_H, \bm{X}'_H) = r(H, \bm{X}_H),
    \end{align}
    thus proving one direction of the proposition. 
    
    For the other direction, notice that a maximally expressive message passing GNN is as powerful as the 1-WL test~\citep{xu2018powerful}, and strictly stronger than a universal set encoder (regarding the set of node features).\footnote{Indeed, a message passing GNN with a maximally expressive pooling layer and no message passing layers is equivalent to a deep set, the latter having proved to be a universal set encoder by~\citet{zaheer2017deep}.} Therefore, by construction~\eqref{eq:construction}, $r(G,\bm{X}_G)=r(H,\bm{X}_H)$ implies that $\exists \bm{P}\in S_n$, $\bm{X}'_G = \bm{PX}'_H$, where $\bm{X}'_G$ is defined in equation~\eqref{eq:_xg_prime_def} but $\bm{X}'_H$ should be alternatively defined as
    \begin{align}
        \bm{X}'_H = \mathrm{concat}\left[\bm{X}_H, g\left(\left\{\mathrm{concat}\left[\mu'_j\mathbf{1}_n, \lambda'_j\mathbf{1}_n, f_{\mu'_j}(\bm{V}'_j)\right]\right\}_{j=1}^{K'}\right)\right],
    \end{align}
    since we have not yet proved that $G$ and $H$ share spectra. The above fact further translates into $\bm{X}_G=\bm{PX}_H$ and
    \begin{align}
        \notag &g\left(\left\{\mathrm{concat}\left[\mu_j\mathbf{1}_n, \lambda_j\mathbf{1}_n, f_{\mu_j}(\bm{V}_j)\right]\right\}_{j=1}^K\right) \\= &\bm{P}g\left(\left\{\mathrm{concat}\left[\mu'_j\mathbf{1}_n, \lambda'_j\mathbf{1}_n, f_{\mu'_j}(\bm{V}'_j)\right]\right\}_{j=1}^{K'}\right).
    \end{align}
    Since $g$ is a universal set representation, the sets on both sides are equal up to an element-wise application of $\bm{P}$. As a consequence,
    \begin{align}
        \{(\mu_j, \lambda_j)\}_{j=1}^K = \{(\mu'_j, \lambda'_j)\}_{j=1}^{K'},
    \end{align}
    or $\mathrm{Spec}\ G = \mathrm{Spec}\ H$. Now that $G$ and $H$ share spectra, we may assume that the eigenvalues $\{\lambda_j\}_{j=1}^K$ are in an order such that $0=\lambda_1<\lambda_2<\cdots<\lambda_K$. We then arrive at equation~\eqref{eq:_concat_equal}, and subsequently $f_{\mu_j}(\bm{V}_j)=\bm{P}f_{\mu_j}(\bm{V}'_j)$, for each $j=1,\ldots, K$. Due to $f_{\mu_j}$ being an $O(\mu_j)$-invariant universal representation, we end up finding that $\exists \bm{Q}_j\in O(\mu_j)$ ($j=1,\ldots, K$), such that $\bm{V}_j = \bm{PV}'_j\bm{Q}_j$. So far we have proved the other direction of the proposition.
\end{proof}

By Proposition~\ref{prop:construction}, the problem of finding a universal representation of graphs is completely reduced to that of finding $O(p)$-invariant universal representations of point sets (as constructions for other components are already known). Therefore, directly applying existing point cloud networks (such as those we will mention in Section~\ref{sect:rel_work}) to graph Laplacian eigenspaces following equation~\eqref{eq:construction} immediately results in a fairly large design space of GNNs, and universality of the resulting GNN directly follows from universality of the underlying point cloud network. 

One may find that equation~\eqref{eq:construction} takes the form of a node feature augmented MPNN. The observation is made explicit with the following definition.

\begin{definition}[Vanilla OGE-Aug]\label{def:vanilla-oge}
    Let $f_p$ be an $O(p)$-invariant universal representation, for each $p=1,2,\ldots$, and $g:2^{\mathbb{R}^3}\rightarrow\mathbb{R}$ be a universal set representation. Define $Z:\mathcal{G}\rightarrow \bigcup_{n=1}^\infty\mathbb{R}^n$ as
    \begin{align}
        Z(G)=g\left(\left\{\mathrm{concat}\left[\mu_j\mathbf{1}_{n(G)}, \lambda_j\mathbf{1}_{n(G)}, f_{\mu_j}(\bm{V}_j)\right]\right\}_{j=1}^K\right),\label{eq:vannila-oge-g}
    \end{align}
    in which the notations follow Proposition~\ref{prop:construction}, and $n(G)\equiv |\mathcal{V}(G)|$. For $G\in\mathcal{G}$, $Z(G)$ is called a \textbf{vanilla orthogonal group equivariant augmentation}, or \textbf{Vanilla OGE-Aug} on $G$. 
\end{definition}

We end this section by discussing the complexity of computing $Z(G)$. The typical complexity of a universal point cloud network is $n \exp(\tilde{O}(\mathrm{dim}))$\footnote{$\tilde{O}(f(n))$ means a complexity linear in $f(n)$ if ignoring poly-logarithm factors, i.e., $O(\log^k f(n))$.}, where $\mathrm{dim}$ is the coordinate dimension. Thus, the complexity of computing Vanilla OGE-Aug is $n\exp(\tilde{O}(\max_{j}\mu_j))$. Our worst-case complexity (in which $\max_j\mu_j\sim n$) matches that of a typical algorithm for graph isomorphism problem (GI). Nevertheless, real-world graphs usually have $\max_j\mu_j\ll n$, making our method computationally affordable in general.

\section{Incorporating the stability constraint}\label{sect:stable}

Proposition~\ref{prop:construction} has theoretically confirmed the possibility of finding universal graph representations with Laplacian eigenvectors, even when the backbone GNN is a (relatively weak) MPNN. Nevertheless, naively applying the network architecture proposed in Proposition~\ref{prop:construction} (or Vanilla OGE-Aug) may not necessarily bring performance gain, due to one important weakness---\textbf{instability}. As is mentioned in Section~\ref{sect:intro}, instability refers to the proneness to produce very different outputs as the input undergoes small perturbations. Instability of Vanilla OGE-Aug stems from the fact that it \textbf{treats Laplacian eigenspaces of different dimensions separately}. It is well known that small perturbations on a graph often makes a high-dimensional (i.e., highly degenerate) eigenspace, say the one spanned by columns of $\bm{U}\in\mathbb{R}^{n\times K}$, split into several low-dimensional ones, say those spanned by columns of $\bm{U}_1\in\mathbb{R}^{n\times k_1},\ldots, \bm{U}_\ell\in\mathbb{R}^{n\times k_\ell}$ respectively. In this case, the eigenvectors $\bm{U}$ undergo a sudden change of encoder after the perturbation (i.e., from $f_K$ to $f_{k_1},\ldots, f_{k_\ell}$, which can be very different). Therefore, the output can vary a lot even if the changes in $\bm{L}$ are small.\footnote{We remark that a similar problem pertains to BasisNet~\citep{lim2022sign}. See the discussion in Appendix C of~\citet{huang2023stability}.} An important lesson from the above discussion is that a ``hard split'' of Laplacian eigenvectors into separate eigenspaces can be susceptible to perturbations, which must be avoided for the sake of stability.

According to equation~\eqref{eq:vannila-oge-g}, in Vanilla OGE-Aug there are two occurrences of explicit dependencies on eigenspace dimensions $\mu_j$ ($j=1,\ldots, K$), namely (i) $\mu_j$ being concatenated as a number, and (ii) a different $f_{\mu_j}$ being used for each value of $\mu_j$. To maintain stability, such dependencies should either be removed, or be replaced by functions not sensitive to the exact eigenspace splitting. Our attempt towards this goal is as follows.

\begin{definition}[OGE-Aug]\label{def:oge-aug}
    Let $G$ be a graph with $n$ nodes. Let $f$ be an $O(n)$-invariant universal representation function. Define
    \begin{align}
        \notag \bm{V}_j^{\text{smooth}}=&\mathrm{concat}[\bm{V}_1\rho(|\lambda_1-\lambda_j|),\\
        &\bm{V}_2\rho(|\lambda_2-\lambda_j|),\ldots, \bm{V}_K\rho(|\lambda_K-\lambda_j|)],
    \end{align}
    where $\rho:\mathbb{R}_{\geqslant 0}\rightarrow [0,1]$ is a continuous \emph{smoothing function} with $\rho(0)=1$ and $\lim_{x\rightarrow+\infty}\rho(x)=0$, and other notations follow Proposition~\ref{prop:construction}. Further let $\phi:\mathbb{R}^2\rightarrow \mathbb{R}^m$ and $\psi:\mathbb{R}^m\rightarrow\mathbb{R}$ be parameterized functions that apply row-wise on $n\times 2$ and $n\times m$ matrices, respectively. Then 
    \begin{align}\label{eq:smooth1}
        Z(G)=\psi\left(\sum_{j=1}^K\mu_j\phi\left(\mathrm{concat}\left[ \lambda_j\mathbf{1}_{n}, f(\bm{V}_j^\text{smooth})\right]\right)\right)
    \end{align}
    is called an \textbf{orthogonal group equivariant augmentation}, or \textbf{OGE-Aug} on $G$.
\end{definition}

\begin{figure*}[t]
    \centering
    \includegraphics[width=0.8\linewidth]{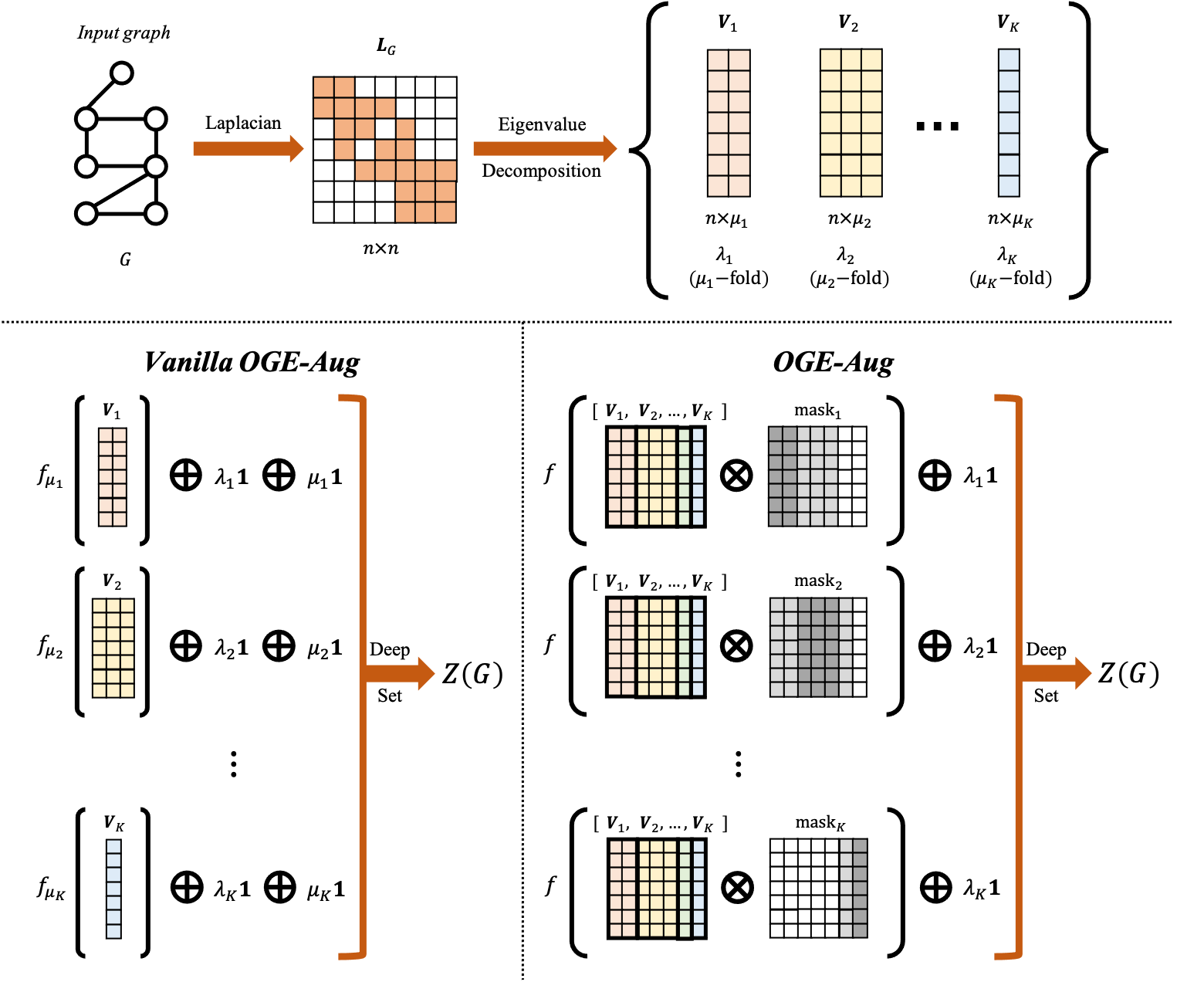}
    \caption{Comparison between Vanilla OGE-Aug and OGE-Aug. Here $\oplus$ represents concatenation, and $\otimes$ represents element-wise multiplication.}
    \label{fig:oge-aug}
\end{figure*}

We illustrate the key differences between Vanilla OGE-Aug and OGE-Aug in Figure~\ref{fig:oge-aug}. We use different colors to represent Laplacian eigenvectors that correspond to different eigenvalues, namely $\bm{V}_1,\bm{V}_2,\ldots, \bm{V}_K$. As is shown on the bottom left of Figure~\ref{fig:oge-aug}, Vanilla OGE-Aug uses a separate $O(\mu)$-invariant universal representation function $f_\mu$ to encode Laplacian eigenvectors with multiplicity $\mu$. In contrast, one may witness from the bottom right of Figure~\ref{fig:oge-aug} that OGE-Aug uses a \textbf{single} $O(n)$-invariant encoder $f$ to encode eigenvectors coming from all eigenspaces, while a continuous smoothing function $\rho$ is used to generate $\mathrm{mask}_1,\mathrm{mask}_2,\ldots,\mathrm{mask}_K$, which keep each group of eigenvectors aware of the eigenspace where they belong, \textbf{as well as the eigenspaces nearby}. 

We remark that OGE-Aug eliminates both sources of instability in Vanilla OGE-Aug mentioned above. First, OGE-Aug avoids using a different encoder $f_{\mu_j}$ for each eigenspace dimension $\mu_j$. Second, the numbers $\mu_j$ ($j=1,\ldots, K$) only appear in the form of a weighted sum, which is insensitive to the exact splitting of Laplacian eigenspaces. Furthermore, with the smoothing function $\rho$, OGE-Aug even gains better flexibility compared with Vanilla OGE-Aug. As $\rho$ becomes more centered at 0 (namely, $\rho(0)=1$ and $\rho(x)\rightarrow 0$ for all $x>0$), each eigenspace gets encoded by its own portion of parameters from $f$ that are not shared with each other; contrarily, with $\rho$ being flatter, more parameters are shared across eigenspaces. In other words, the shape of $\rho$ controls the ``degree of smoothness'' of OGE-Aug.

Next, we quantitatively characterize the stability of OGE-Aug. To this end, we first present our definition of stability, following (though slightly different from)~\citep{huang2023stability}.

\begin{definition}[Stability, following Definition 3.1 of \citep{huang2023stability}]
    A function $f$, operating on the Laplacian $\bm{L}$ of a graph $G$ and producing a node feature augmentation $\bm{Z}\in\mathbb{R}^{|\mathcal{V}(G)|\times d}$, is said to be stable, if there exist constants $c_1, C_1,\ldots, c_m,C_m>0$, such that for any two Laplacians $\bm{L},\bm{L}'$,
    \begin{align}
        \notag \|f(\bm{L})-\bm{P}_*f(\bm{L}')\|_\text{F}&\leqslant \max 
        \big\{C_1\cdot \|\bm{L}-\bm{P}_*\bm{L}'\bm{P}_*^T\|_\text{F}^{c_1},\\
        &\ldots, C_m\cdot \|\bm{L}-\bm{P}_*\bm{L}'\bm{P}_*^T\|_\text{F}^{c_m}\big\}
    \end{align}
    in which $\|\cdot\|_\text{F}$ stands for Frobenius norm, and $\bm{P}_*=\arg\min_{\bm{P}\in S_n}\|\bm{L}-\bm{PL}'\bm{P}^T\|_\text{F}$ is the permutation matrix matching $\bm{L}$ and $\bm{L}'$ (assuming both $\bm{L}$ and $\bm{L}'$ are of size $n\times n$).
\end{definition}

We are now ready to give our theoretical result on the stability of OGE-Aug. We assume that the following conditions hold for functions $\psi, \phi, f$ and $\rho$.

\begin{enumerate}
    \item $\psi, \phi$ and $\rho$ are Lipschitz continuous, with Lipschitz constants $J_\psi, J_\phi$ and $J_\rho$ respectively. Namely, $\|\psi(\bm{X})-\psi(\bm{X}')\|_\text{F}\leqslant J_\psi \|\bm{X}-\bm{X}'\|_\text{F}$ for any $\bm{X}, \bm{X}'\in\mathbb{R}^{n\times m}$, $\|\phi(\bm{X})-\phi(\bm{X}')\|_\text{F}\leqslant J_\phi \|\bm{X}-\bm{X}'\|_\text{F}$ for any $\bm{X},\bm{X}'\in\mathbb{R}^{n\times 2}$, and $|\rho(x)-\rho(x')|\leqslant J_\rho |x-x'|$ for any $x,x'\in\mathbb{R}_{\geqslant 0}$.
    \item $f$ satisfies the following condition: $\exists J_f>0$, such that for any $\bm{X}, \bm{X}'\in\mathbb{R}^{n\times n}$, holds $\|f(\bm{X})-f(\bm{X}')\|\leqslant J_f \min_{\bm{Q}\in O(n)}\|\bm{X}-\bm{X}'\bm{Q}\|_\text{F}$. One may see $f$ as $J_f$-Lipschitz continuous after rotating its arguments along the same direction.
    \item There exists a constant $\delta>0$, such that $\rho(x)=0$ for all $x>\delta$.
\end{enumerate}

Given the above assumptions, we have

\begin{proposition}[Stability of OGE-Aug]\label{prop:stability_bound}
    With the assumptions on $\psi, \phi, f$ and $\rho$ specified above, OGE-Aug defined by \eqref{eq:smooth1} is stable. To be specific, given two graphs $G, G'\in\mathcal{G}$ with Laplacians $\bm{L}$ and $\bm{L}'$ respectively, there exists a proper value of $\delta$ such that
        \begin{align}
        \notag\|Z(G)-\bm{P}_*Z(G')\|_\text{F} &\leqslant nJ_\psi J_\phi\big[(\sqrt{n}+2nJ_\rho J_f)\|\bm{L}-\bm{P}_*\bm{L}'\bm{P}_*^T\|_2\\
        &\ +4\sqrt[4]{2}J_f\sqrt{J_\rho}|\mathcal{J}|_\text{max}\|\bm{L}-\bm{P}_*\bm{L}'\bm{P}_*^T\|_\text{F}^{1/2}\big],\label{eq:stability_final}
    \end{align}
    where $\|\cdot \|_2$ is the spectral norm which is no larger than the Frobenius norm $\|\cdot\|_\text{F}$, and $n=|\mathcal{V}(G)|=|\mathcal{V}(G')|$. $|\mathcal{J}|_\text{max}$ is defined as following. Let $\tilde{\lambda}'_1\leqslant \tilde{\lambda}'_2\leqslant \cdots\leqslant \tilde{\lambda}'_n$ be the $n$ eigenvalues of $\bm{L}'$, and $\mathcal{J}=\{\tilde{\lambda}'_{j+1},\ldots, \tilde{\lambda}'_k\}$ be a continuous range of eigenvalues such that $\tilde{\lambda}'_{j+1}-\tilde{\lambda}'_j>\delta$ and $\tilde{\lambda}'_{k+1}-\tilde{\lambda}'_k>\delta$ yet $\tilde{\lambda}'_{m+1}-\tilde{\lambda}'_m\leqslant \delta$ for each $m=j+1,j+2,\ldots, k-1$ (where we define $\tilde{\lambda}'_0=-\infty$ and $\tilde{\lambda}'_{n+1}=+\infty$ for convenience). $|\mathcal{J}|_\text{max}$ is the maximum possible size of such $\mathcal{J}$.
\end{proposition}

We give the proof in Appendix~\ref{app:proof-stable}. To ensure that the inequality~\eqref{eq:stability_final} holds, in principle we need to tune $\delta$ for different $G$ and $G'$. However, in our experiments we simply take $\delta$ as a hyperparameter designated before actual training.

Finally, we discuss practical implementations of OGE-Aug. While presenting the universality result (Proposition~\ref{prop:construction}), we have assumed that $f_p$ ($p=1,2,\ldots$) is an $O(p)$-invariant universal representation function. This universality requirement is inherited to OGE-Aug (Definition~\ref{def:oge-aug}). Namely, we still require that $f$ is an $O(n)$-invariant universal representation. We now point out that such requirement, despite producing maximally expressive networks in theory, can be impractical to implement. A straightforward reason is that with $f$ being universal, the resulting network architecture has a typical complexity of $n\exp(\tilde{O}(n))$ which is generally unacceptable. Additionally, insisting on the universality of $f$ can be harmful to the stability of OGE-Aug, since a more expressive $f$ might result in a larger Lipschitz constant $J_f$. Therefore, in our actual implementation of OGE-Aug, we no longer require $f$ to be universal. Instead, we adopt as $f$ a Cartesian tensor based point cloud network~\citep{finkelshtein2022simple} with Cartesian tensors up to the second order used. We include more experimental details, as well as a complexity analysis for our implementation, in Appendix~\ref{Sec:appendix_experiment}.

\section{Related works}\label{sect:rel_work}

\textbf{Graph representation learning with Laplacian eigenvectors.} It is well-known that eigenvectors of graph Laplacian corresponding to the smallest eigenvalues contain ``positional'' information of nodes. A number of works have thus adopted Laplacian eigenvectors as a technique for node feature augmentation. As we have mentioned in Section~\ref{sect:intro}, there are two important issues regarding the application of Laplacian eigenvectors in graph representation learning, namely orthogonal group invariance (or sign-and-basis invariance) and stability. Some early works~\citep{dwivedi2020generalization, kreuzer2021rethinking} have noticed the sign invariance problem and tried to alleviate it by randomly flipping the signs of Laplacian eigenvectors, while completely ignored the basis invariance problem. \citet{lim2022sign} is the first work to formally state and systematically address the sign-and-basis invariance issue. Nevertheless, it fails to meet the stronger requirement of stability. So far, only two works~\citep{wang2022equivariant, huang2023stability} have seriously discussed the stability issue by giving mathematical definitions for it, and proposing learning methods that are provably stable.

\begin{table*}[t]
\caption{QM9 results (MAE $\downarrow$). Highlighted are \textbf{\textcolor{red}{first}}, \textbf{second} best results.}
\label{Table_QM9}
\tabcolsep=0.05cm
\begin{tabular}{l|ccccccccc|c}
\toprule
Target & MPNN & 1-2-3-GNN & DTNN & DeepLRP & PPGN & NGNN & KP-GIN+ & 4-IDMPNN & PST & OGE-Aug\\
\midrule
$\mu$ & 0.358 & 0.476 & 0.244 & 0.364 & \textbf{0.231} & 0.433 & 0.358 & 0.398 & 0.319 & \textbf{\textcolor{red}{0.0822}} \\
$\alpha$ & 0.89 & 0.27 & 0.95 & 0.298 & 0.382 & 0.265 & 0.233 & 0.226 & \textbf{0.189} & \textbf{\textcolor{red}{0.159}} \\
$\varepsilon_{\textrm{HOMO}}$ & 0.00541 &  0.00337 & 0.00388 & 0.00254 & 0.00276 & 0.00279 & 0.00240 & 0.00263 & \textbf{0.00220} & \textbf{\textcolor{red}{0.00140}}\\
$\varepsilon_{\textrm{LUMO}}$ & 0.00623 &  0.00351 & 0.00512 & 0.00277 & 0.00287 & 0.00276 & 0.00236 & 0.00286 & \textbf{0.00215} & \textbf{\textcolor{red}{0.00144}}\\
$\Delta \varepsilon$  & 0.0066 &  0.0048& 0.0112 & 0.00353 & 0.00406 & 0.00390 & 0.00333 & 0.00398 & \textbf{0.00311} & \textbf{\textcolor{red}{0.00198}}\\
$\langle R^2 \rangle$ & 28.5 &  22.9 & 17.0 & 19.3 & 16.7 & 20.1 & 16.49 & \textbf{10.4} & 13.08 & \textbf{\textcolor{red}{5.55}}\\
$\textrm{ZPVE}$ & 0.00216 &  0.00019 & 0.00172 & 0.00055 & 0.00064 & 0.00015 & 0.00017 & \textbf{\textcolor{red}{0.00013}} & \textbf{0.00014} & 0.000149\\
$U_0$ & 2.05 & \textbf{0.0427} & 2.43 & 0.413 & 0.234 & 0.205 & 0.0682 & \textbf{\textcolor{red}{0.0189}} & 0.127 & 0.0526\\
$U$ & 2.00 & 0.111 & 2.43 & 0.413 & 0.234 & 0.200 & 0.0553 & \textbf{\textcolor{red}{0.0152}} & 0.130 & \textbf{0.0356}\\
$H$ & 2.02 & \textbf{0.0419} & 2.43 & 0.413 & 0.229 & 0.249 & 0.0575 & \textbf{\textcolor{red}{0.0160}} & 0.128 & 0.0439\\
$G$ & 2.02 & 0.0469 & 2.43 & 0.413 & 0.238 & 0.253 & 0.0484 & \textbf{\textcolor{red}{0.0159}} & 0.130 & \textbf{0.0441}\\
$c_{\textrm{v}}$ & 0.42 &  0.0944 & 0.27 & 0.129 & 0.184 & 0.0811 & 0.0869 & 0.0890 & \textbf{0.0777} & \textbf{\textcolor{red}{0.0681}}\\

\bottomrule
\end{tabular}
\end{table*}

\textbf{Orthogonal-group invariant networks.} Orthogonal-group invariant neural networks are those taking as input one or more $p$-dimensional vector(s) and outputing an $O(p)$-invariant scalar, i.e., a value invariant with respect to an $O(p)$ transformation on the input vector system. As is pointed out by, e.g.,~\citet{bronstein2021geometric}, orthogonal-group invariance is a desirable property for learning tasks on molecular data or point clouds, in which Euclidean coordinates play important roles. 

Orthogonal-group equivariance is a property closely related to invariance. An $O(p)$-equivariant network takes as input one or more representation(s)\footnote{In our context, a representation of $O(p)$ means a vector lying in a linear space $\mathcal{L}$, given that a group homomorphism from $O(p)$ to the general linear group $\mathrm{GL}(\mathcal{L})$ on $\mathcal{L}$ exists.} of $O(p)$ (with $p$-dimensional vectors being a special case), and outputs another (or another set of) representation(s) of $O(p)$, in a way that whenever the input system undergoes the action of an $O(p)$ group element $g$, the output also undergoes an action corresponding to $g$. In practice, invariant networks are usually constructed by stacking multiple equivariant layers, along with a final invariant layer. Regarding the intermediate orthogonal group representations they use, existing works on the design of invariant networks mainly take one of the four approaches: (i) utilizing scalar or vector representations~\citep{deng2021vector, li2024distance, satorras2021n, villar2021scalars}; (ii) utilizing hand-crafted higher-order representations~\citep{gasteiger2020directional, gasteiger2021gemnet, schutt2021equivariant}; (iii) utilizing higher-order Cartesian tensor representations~\citep{finkelshtein2022simple, ruhe2024clifford}; (iv) utilizing higher-order irreducible representations~\citep{batzner20223, bogatskiy2020lorentz, cohen2018spherical, fuchs2020se, thomas2018tensor}.

Similar to the question of expressive power of GNNs, there exists the question of whether an orthogonal-group invariant network can express all possible geometric configurations (either of a single vector or of a point cloud) up to an arbitrary orthogonal transformation. Invariant networks possessing the above property are usually called \emph{universal}. Several works have established universality of some architectures mentioned above. \citet{villar2021scalars} shows that universality can be achieved merely using scalar and vector representations, as long as interaction terms including sufficiently many vectors are allowed, and that the network output is restricted to be scalars or vectors. \citet{li2024distance} further shows by construction that an invariant network can be already universal with 4-vector interaction terms, even if all intermediate representations are restricted scalar. Regarding methods using higher-order representations, \citet{dym2020universality} proves the universality of two specific architectures exploiting higher-order irreducible representations of SO(3)---Tensor Field Networks (TFN)~\citep{thomas2018tensor} and SE(3)-Transformers~\citep{fuchs2020se}. Based on TFN, \citet{finkelshtein2022simple} proposes another universal architecture utilizing Cartesian tensor representations. The universality results reviewed above have laid theoretical foundations for our proposed method.

\textbf{Expressive GNNs.} As is shown by \citet{xu2018powerful}, the expressive power of MPNNs is upper-bounded by that of 1-dimensional Weisfeiler-Leman test (1-WL). This implies that MPNNs can fail to discriminate many non-isomorphic graph pairs, potentially leading to their weakness in capturing important structural information or multi-node interactions. A great number of works have attempted to improve the expressive power of GNNs, in the sense that to make them better either at solving the graph isomorphism problem (GI), or at approximating certain graph functions. Those existing works can be roughly categorized into three families: (1) methods utilizing additional combinatorial features~\citep{barcelo2021graph, bouritsas2022improving, li2020distance}; (2) methods applying message passing among higher-order tuples of nodes, or \emph{higher-order GNNs}~\citep{bodnar2021weisfeiler, feng2023towards, maron2018invariant, maron2019provably, morris2019weisfeiler, morris2020weisfeiler, zhang2023complete, pmlr-v202-zhou23n, zhou2023distance}; (3) methods decomposing input graphs into bags of subgraphs, or \emph{subgraph GNNs}~\citep{bevilacqua2024efficient, cotta2021reconstruction, frascaunderstanding2022, huang2023boosting, kong2023maggnn, qianordered2022, you2021identity, zhang2021nested, pmlr-v202-zhou23n}. While methods belonging to class (1) enjoy the lowest complexities, they often generalize worse due to their use of hand-crafted features. On the contrary, higher-order GNNs and subgraph GNNs bring more systematic gains to the expressive power, but their computational complexities are much higher than MPNNs. Hence, a trade-off between expressive power and efficiency is an important issue for the design of expressive GNNs.

\textbf{Graph transformers.} Graph transformers~\citep{chen2022structure, dwivedi2021graph, rampavsek2022recipe, wang2024graph, ying2021transformers} treat each node within a graph as a separate token, and use a standard transformer architecture to update node features (or embeddings of tokens). With attention mechanism, graph transformers take into account the interactions between all pairs of nodes (instead of only connected node pairs, as in traditional MPNNs), and are naturally good at capturing long-range interactions~\citep{dwivedi2022long}. One of the central issues regarding graph transformers is the design of structural and positional encodings of nodes, in order to make transformers aware of adjacency information. \citet{Kim2023pure, pmlr-v238-zhou24a} analyze the theoretical expressive power of graph transformers and their high-order versions as well as the effects of positional encodings.

\section{Experiments}\label{sect:exp1}

In this section, we conduct extensive experiments to evaluate the performance of our methods. A number of graph datasets, including synthetic and real-world ones, are adopted. We present in the main paper results on three real-world datasets: QM9~\citep{QM9}, ZINC12k~\citep{Zinc} and PCQM-Contact~\citep{dwivedi2022long}, while leaving other experimental results to Appendix~\ref{Sec:appendix_experiment}. Dataset statistics and detailed experimental settings can also be found in Appendix~\ref{Sec:appendix_experiment}.



\textbf{QM9.}
QM9~\citep{QM9} is a graph property regression dataset containing 130k small molecules and 19 regression targets. We use a commonly adopted 0.8/0.1/0.1 training/validation/test split ratio, and report the results of the first 12 targets.
Several representative expressive GNNs are selected as baselines, including MPNN, 1-2-3-GNN~\citep{morris2019weisfeiler}, DTNN~\citep{DTNN}, DeepLRP~\citep{chen2020can}, PPGN~\citep{maron2019provably}, NGNN~\citep{zhang2021nested}, KP-GIN+~\citep{Khop}, IDMPNN~\citep{pmlr-v202-zhou23n} and PST~\citep{wang2024graph}. To ensure a fair comparison, methods utilizing geometric coordinates of atoms are not included. The results are shown in Table~\ref{Table_QM9}. From Table~\ref{Table_QM9}, we find that OGE-Aug achieves competitive performance on all 12 targets. We also notice that our method achieves a relatively low MAE on targets $U_0$, $U$, $H$ and $G$, compared with subtree- or subgraph-based methods such as MPNN, NGNN or KP-GIN+, as well as other Laplacian eigenvector augmented GNNs like PST. This fact indicates that our method has the ability to capture \textbf{global properties} of graphs, since those targets are macroscopic thermodynamic properties of molecules and heavily depend on long-range interactions (for example, intermolecular forces like hydrogen bonds).

\begin{table}[t]
    \caption{ZINC12k results (MAE $\downarrow$). Shown is the mean $\pm$ std of 5 runs.}
\vspace{-2pt}
\label{Table_zinc}
\begin{center}
\begin{tabular}{l c}
\toprule
Method & Test MAE \\
\midrule
GIN  & $0.163\pm0.004$\\
KP-GIN+ & $0.119\pm0.002$\\
GNN-AK+ & $0.080\pm0.001$\\
CIN & $0.079\pm0.006$\\
\midrule
GIN, with PEG & $0.144\pm0.008$\\
GIN, with SignNet & $0.085\pm0.003$\\
GIN, with BasisNet & $0.155\pm0.007$\\
GIN, with SPE & $0.069\pm 0.004$\\
\midrule
SAN & $0.139 \pm 0.006$\\
Graphormer & $0.122 \pm 0.006$\\
GPS & $0.070\pm0.004$\\
Specformer & $0.066 \pm 0.003$\\
\midrule
GINE, with OGE-Aug & $0.066\pm 0.002$\\
GPS, with OGE-Aug & \textbf{0.064 $\pm$ 0.003}\\
\bottomrule
\end{tabular}
\end{center}
\vspace{-5pt}
\end{table}

\begin{table}
    \caption{Results on PCQM-Contact dataset from the long-range graph benchmarks (LRGB). Highlighted are the \textcolor{orange}{first}, \textcolor{teal}{second}, \textcolor{violet}{third} best results.}
\label{LRGB}
\begin{center}
\begin{tabular}{lcc}
\toprule
model & PE & MRR $\uparrow$ \\
\midrule
GCN & None & $0.3234\pm 0.0006$ \\
GINE & None & $0.3180\pm 0.0027$ \\
GatedGCN & None & $0.3218\pm 0.0011$ \\
Transformer & LapPE & $0.3174\pm 0.0020$ \\
SAN & LapPE & $0.3350 \pm 0.0003$ \\
SAN & RWSE & $0.3341 \pm 0.0006$ \\
GPS & LapPE & $0.3337\pm 0.0006$\\
GPS & EdgeRWSE & \textbf{\textcolor{teal}{0.3408 $\pm$ 0.0003}} \\
GPS & Hodge1Lap & \textbf{\textcolor{violet}{0.3407 $\pm$ 0.0004}} \\
\midrule
GPS & OGE-Aug  & \textbf{\textcolor{orange}{0.3543 $\pm$ 0.0004}}\\
\bottomrule
\end{tabular}
\end{center}
\vspace{-2pt}
\end{table}

     \textbf{ZINC.} ZINC12k~\citep{Zinc} is a subset of the ZINC250k dataset containing 12k molecules, and the task is molecular property (constrained solubility) regression evaluated by mean absolute error (MAE). We follow the official split of the dataset. We include common baselines such as GIN~\citep{xu2018powerful}, PNA~\citep{PNA}, DeepLRP, OSAN~\citep{qianordered2022}, KP-GIN+, GNN-AK+~\citep{GNNAK} and CIN~\citep{bodnar2021weisfeiler}. \vspace{0.2em}

     We also include previous methods making use of Laplacian eigenvectors to produce node feature augmentations (which are usually named \emph{positional encodings}), such as PEG~\citep{wang2022equivariant}, SignNet~\citep{lim2022sign}, BasisNet~\citep{lim2022sign} and SPE~\citep{huang2023stability}, as well as graph transformers such as SAN~\citep{kreuzer2021rethinking}, Graphormer~\citep{Graphormer}, GraphGPS~\citep{rampavsek2022recipe} and Specformer~\citep{specformer}. Among the graph transformer baselines, SAN, GraphGPS and Specformer also encode spectral information through other approaches. Regarding our OGE-Aug, we consider both GINE~\citep{PretrainGINE} (which belongs to the MPNN family) and the GPS as base models. As shown in Table~\ref{Table_zinc}, OGE-Aug outperforms all baseline methods even combined with the simple GINE backbone without global attention.\vspace{0.2em}

     \textbf{PCQM-Contact.} As part of the long-range graph benchmarks (LRGB)~\citep{dwivedi2022long}, PCQM-Contact is a dataset derived from PCQM4Mv2. The task is binary link ranking measured by the Mean Reciprocal Rank (MRR), which requires the ability to capture long-range interactions. MPNN baselines include GCN~\citep{kipf2016semi}, GINE~\citep{PretrainGINE}, and GatedGCN~\citep{ResidualGatedGCN}, while graph transformer baselines include Transformer, SAN and GPS combined with positional encodings (PEs) like LapPE~\citep{kreuzer2021rethinking}, RWSE~\citep{dwivedi2021graph}, EdgeRWSE~\citep{HodgeRandomWalk} and Hodge1Lap~\citep{HodgeRandomWalk}. We combine GPS with our OGE-Aug and achieve best performance across all baselines, which verifies the benefit of bringing in long-range information via OGE-Aug.

\textbf{Other graph benchmarks.} We further evaluate the performance of OGE-Aug on three additional graph learning benchmarks: CLUSTER~\citep{dwivedi2023benchmarking}, PATTERN~\citep{dwivedi2023benchmarking} and ogbg-molhiv~\citep{hu2021opengraphbenchmarkdatasets}. CLUSTER and PATTERN are node classification datasets, while ogbg-molhiv is a graph classification dataset. The results are summarized in Table~\ref{Table_other_benchmark}. We quote the baseline results directly from~\citet{rampavsek2022recipe} and~\citet{shirzad2023exphormersparsetransformersgraphs}. One may find that OGE-Aug outperforms all baselines on the three datasets.

\begin{table}[t]
    \caption{Five-run results on CLUSTER, PATTERN and ogbg-molhiv.}
\label{Table_other_benchmark}
\vspace{-5pt}
\begin{center}
\setlength{\tabcolsep}{1mm}
\begin{tabular}{lccc}
\toprule
Method & CLUSTER  & PATTERN & ogbg-molhiv  \\
\midrule
Metric& Acc $\uparrow$ & Acc $\uparrow$ & AUROC $\uparrow$\\
\midrule
GCN            & 68.50 $\pm$ 0.98        & 71.89 $\pm$ 0.34         & 75.99 $\pm$ 1.19\\
GIN            & 64.72 $\pm$ 1.55        & 85.39 $\pm$ 0.14         & 77.07 $\pm$ 1.49\\
GAT            & 70.59 $\pm$ 0.45        & 78.27 $\pm$ 0.19         & -                \\
GatedGCN       & 73.84 $\pm$ 0.33        & 85.57 $\pm$ 0.09         & 78.74 $\pm$ 1.19\\
SAN            & 76.69 $\pm$ 0.65          & 86.58 $\pm$ 0.37           & 77.85 $\pm$ 2.47\\
K-Subgraph SAT & 77.86 $\pm$ 0.10          & 86.85 $\pm$ 0.37           & -              \\
GraphGPS       & 78.02 $\pm$ 0.18          & 86.69 $\pm$ 0.59           & 78.80 $\pm$ 1.01\\
Exphormer      & 78.07 $\pm$ 0.04        & 86.74 $\pm$ 0.15           & -             \\
OGE-Aug        & \textbf{78.33 $\pm$ 0.13}      & \textbf{86.87 $\pm$ 0.33}    &\textbf{80.01 $\pm$ 0.59}\\
\bottomrule
\end{tabular}
\end{center}
\vspace{-12pt}
\end{table}

\textbf{OOD benchmarks.} We evaluate the OOD performance of OGE-Aug on DrugOOD~\citep{ji2022drugoodoutofdistributionooddataset}, an OOD benchmark for drug discovery. We consider three domains on which distribution shifts exist, namely Assay (which assay the molecule belongs to), Scaffold (core structure of the molecule) and Size (size of the molecule). For each domain, the dataset is divided into five splits: the training set, the in-distribution (ID) validation/test sets, and the out-of-distribution (OOD) validation/test sets. The data distribution of OOD splits is different from that of ID splits regarding the specific domain. The task is graph-level binary classification, i.e., to predict whether the drug is active. We use AUROC as the evaluation metric.

The experimental results are shown in Table~\ref{Table_DrugOOD}. We choose PE methods from~\citep{huang2023stability} as our baselines. Our OGE-Aug outperforms all baselines on the Assay domain, and achieves comparable results on Scaffold and Size domains. Moreover, the performance of our method is better than that of BasisNet on 5 out of the 6 OOD evaluation targets, verifying the benefits of possessing theoretically guaranteed stability.

\begin{table}[t]
    \caption{AUROC (the larger, the better) results on DrugOOD.}
\label{Table_DrugOOD}
\vspace{-5pt}
\begin{center}
\setlength{\tabcolsep}{1mm}
\begin{tabular}{llcccc}
\toprule
\multirow{2}{*}{Domain} & \multirow{2}{*}{Method} & ID-Val & ID-Test  & OOD-Val & OOD-Test \\
&&(AUROC)&(AUROC)&(AUROC)&(AUROC)\\
\midrule
\multirow{6}{*}{Assay} & No PE     & 92.92           & 92.89          & 71.02          & 71.68     \\
          & PEG       & 92.51           & 92.57          & 70.86          & 71.98          \\
                    & SignNet   & 92.26           & 92.43          & 70.16          & 72.27          \\
          & BasisNet  & 88.96           & 89.42          & 71.19          & 71.66          \\
          & SPE       & 92.84           & \textbf{92.94}      & 71.26          & 72.53          \\
          & OGE-Aug   & \textbf{94.88}       & 86.75          & \textbf{82.26}      & \textbf{73.73}      \\
          \midrule
\multirow{6}{*}{Scaffold} & No PE     & 96.56           & 87.95          & 79.07          & 68.00          \\
          & PEG       & 95.65           & 86.20          & 79.17          & 69.15          \\
          & SignNet   & 95.48           & 86.73          & 77.81          & 66.43          \\
          & BasisNet  & 85.80           & 78.44          & 73.36          & 66.32          \\
          & SPE       & \textbf{96.32}       & \textbf{88.12}      & \textbf{80.03}      & \textbf{69.64}      \\
          & OGE-Aug   & 95.02           & 86.54          & 78.67          & 65.94          \\
          \midrule
 \multirow{6}{*}{Size}     & No PE     & \textbf{93.78}       & \textbf{93.60}      & \textbf{82.76}      & \textbf{66.04}      \\
          & PEG       & 92.46           & 92.67          & 82.12          & 66.01          \\
          & SignNet   & 93.30           & 93.20          & 80.67          & 64.03          \\
          & BasisNet  & 86.04           & 85.51          & 75.97          & 60.79          \\
          & SPE       & 92.46           & 92.67          & 82.12          & 66.02          \\
          & OGE-Aug   & 94.65           & 84.88          & 78.44          & 64.64  \\
\bottomrule
\end{tabular}
\vspace{-5pt}
\end{center}
\end{table}

\textbf{Ablation studies.} We also study the effect of the smoothing function $\rho(\cdot)$ in OGE-Aug. We use ZINC as the evaluation dataset. We take GINE as the base model, and apply either Vanilla OGE-Aug, or OGE-Aug with different smoothing functions $\rho(\cdot)$ (all of them taking the form of equation~\eqref{eq:rho_shape_actual} but with different hyperparameters $\delta$). The results are shown in Table~\ref{Table_ablation}.

We find that applying Vanilla OGE-Aug instead of OGE-Aug leads to significant performance drop, which verifies the importance of ensuring stability by introducing the smoothing function $\rho$. We also observe that as long as the hyperparameter $\delta$ is not too close to zero, the performance varies little with different choices of $\delta$.

\begin{table}[t]
    \caption{Ablation studies on ZINC.}
\label{Table_ablation}
\vspace{-7pt}
\begin{center}
\begin{tabular}{lc}
\toprule
Method & MAE ($\downarrow$)\\
\midrule
Vanilla OGE-Aug & 0.098\\
OGE-Aug ($\delta=5\times 10^{-3}$) &	0.066\\
OGE-Aug ($\delta=5\times 10^{-2}$)	&0.066\\
OGE-Aug ($\delta=5\times 10^{-1}$)	&0.065\\
\bottomrule
\end{tabular}
\end{center}
\vspace{-12pt}
\end{table}

\section{Conclusion}

In this paper, we propose to apply orthogonal group invariant neural networks on Laplacian eigenspaces of graphs, so as to produce node feature augmentations that may possess great expressive power. We present Vanilla OGE-Aug and OGE-Aug as two instances of our proposed framework, of which the former illustrates the potential of our method to achieve universal representation of graphs, while the latter is provably stable and practically useful. Extensive experiments have verified the outstanding performance of OGE-Aug on various benchmarks as well as its capability to learn global properties of graphs. We remark that our approach to incorporating stability into graph learning methods based on Laplacian eigenvectors, i.e., by ensuring \emph{smoothness} while processing different Laplacian eigenspaces, is a general technique, and can be applied to other machine learning domains where eigenvalues and eigenvectors are of significant interest.

\section*{Acknowledgement}

This work is supported by National Natural Science Foundation of China (62550138, 62276003).

\newpage
\bibliographystyle{ACM-Reference-Format}
\bibliography{sample-base}

@String{Computing = "Computing" }

@String{Computer = "{IEEE} Computer" }

@String{Springer = "Springer-Verlag" }

@ArtifactSoftware{R,
    title = {R: A Language and Environment for Statistical Computing},
    author = {{R Core Team}},
    organization = {R Foundation for Statistical Computing},
    address = {Vienna, Austria},
    year = {2019},
    url = {https://www.R-project.org/},
}

@article{dwivedi2023benchmarking,
  title={Benchmarking graph neural networks},
  author={Dwivedi, Vijay Prakash and Joshi, Chaitanya K and Luu, Anh Tuan and Laurent, Thomas and Bengio, Yoshua and Bresson, Xavier},
  journal={Journal of Machine Learning Research},
  volume={24},
  number={43},
  pages={1--48},
  year={2023}
}

@misc{shirzad2023exphormersparsetransformersgraphs,
      title={Exphormer: Sparse Transformers for Graphs}, 
      author={Hamed Shirzad and Ameya Velingker and Balaji Venkatachalam and Danica J. Sutherland and Ali Kemal Sinop},
      year={2023},
      eprint={2303.06147},
      archivePrefix={arXiv},
      primaryClass={cs.LG},
      url={https://arxiv.org/abs/2303.06147}, 
}

@misc{ji2022drugoodoutofdistributionooddataset,
      title={DrugOOD: Out-of-Distribution (OOD) Dataset Curator and Benchmark for AI-aided Drug Discovery -- A Focus on Affinity Prediction Problems with Noise Annotations}, 
      author={Yuanfeng Ji and Lu Zhang and Jiaxiang Wu and Bingzhe Wu and Long-Kai Huang and Tingyang Xu and Yu Rong and Lanqing Li and Jie Ren and Ding Xue and Houtim Lai and Shaoyong Xu and Jing Feng and Wei Liu and Ping Luo and Shuigeng Zhou and Junzhou Huang and Peilin Zhao and Yatao Bian},
      year={2022},
      eprint={2201.09637},
      archivePrefix={arXiv},
      primaryClass={cs.LG},
      url={https://arxiv.org/abs/2201.09637}, 
}

@misc{hu2021opengraphbenchmarkdatasets,
      title={Open Graph Benchmark: Datasets for Machine Learning on Graphs}, 
      author={Weihua Hu and Matthias Fey and Marinka Zitnik and Yuxiao Dong and Hongyu Ren and Bowen Liu and Michele Catasta and Jure Leskovec},
      year={2021},
      eprint={2005.00687},
      archivePrefix={arXiv},
      primaryClass={cs.LG},
      url={https://arxiv.org/abs/2005.00687}, 
}

@article{dwivedi2021graph,
  title={Graph neural networks with learnable structural and positional representations},
  author={Dwivedi, Vijay Prakash and Luu, Anh Tuan and Laurent, Thomas and Bengio, Yoshua and Bresson, Xavier},
  journal={arXiv preprint arXiv:2110.07875},
  year={2021}
}

@article{dwivedi2020generalization,
  title={A generalization of transformer networks to graphs},
  author={Dwivedi, Vijay Prakash and Bresson, Xavier},
  journal={arXiv preprint arXiv:2012.09699},
  year={2020}
}

@article{rampavsek2022recipe,
  title={Recipe for a general, powerful, scalable graph transformer},
  author={Ramp{\'a}{\v{s}}ek, Ladislav and Galkin, Michael and Dwivedi, Vijay Prakash and Luu, Anh Tuan and Wolf, Guy and Beaini, Dominique},
  journal={Advances in Neural Information Processing Systems},
  volume={35},
  pages={14501--14515},
  year={2022}
}

@article{kreuzer2021rethinking,
  title={Rethinking graph transformers with spectral attention},
  author={Kreuzer, Devin and Beaini, Dominique and Hamilton, Will and L{\'e}tourneau, Vincent and Tossou, Prudencio},
  journal={Advances in Neural Information Processing Systems},
  volume={34},
  pages={21618--21629},
  year={2021}
}

@article{lim2022sign,
  title={Sign and basis invariant networks for spectral graph representation learning},
  author={Lim, Derek and Robinson, Joshua and Zhao, Lingxiao and Smidt, Tess and Sra, Suvrit and Maron, Haggai and Jegelka, Stefanie},
  journal={arXiv preprint arXiv:2202.13013},
  year={2022}
}

@inproceedings{huang2023stability,
  title={On the stability of expressive positional encodings for graph neural networks},
  author={Huang, Yinan and Lu, William and Robinson, Joshua and Yang, Yu and Zhang, Muhan and Jegelka, Stefanie and Li, Pan},
  booktitle={The Twelfth International Conference on Learning Representations},
  year={2024}
}

@article{wang2022equivariant,
  title={Equivariant and stable positional encoding for more powerful graph neural networks},
  author={Wang, Haorui and Yin, Haoteng and Zhang, Muhan and Li, Pan},
  journal={arXiv preprint arXiv:2203.00199},
  year={2022}
}

@article{wang2024graph,
  title={Graph as Point Set},
  author={Wang, Xiyuan and Li, Pan and Zhang, Muhan},
  journal={arXiv preprint arXiv:2405.02795},
  year={2024}
}

@inproceedings{xu2018powerful,
  title={How Powerful are Graph Neural Networks?},
  author={Xu, Keyulu and Hu, Weihua and Leskovec, Jure and Jegelka, Stefanie},
  booktitle={International Conference on Learning Representations},
  year={2018}
}

@inproceedings{morris2019weisfeiler,
  title={Weisfeiler and leman go neural: Higher-order graph neural networks},
  author={Morris, Christopher and Ritzert, Martin and Fey, Matthias and Hamilton, William L and Lenssen, Jan Eric and Rattan, Gaurav and Grohe, Martin},
  booktitle={Proceedings of the AAAI conference on artificial intelligence},
  pages={4602--4609},
  year={2019}
}

@article{morris2020weisfeiler,
  title={Weisfeiler and Leman go sparse: Towards scalable higher-order graph embeddings},
  author={Morris, Christopher and Rattan, Gaurav and Mutzel, Petra},
  journal={Advances in Neural Information Processing Systems},
  volume={33},
  pages={21824--21840},
  year={2020}
}

@inproceedings{you2021identity,
  title={Identity-aware graph neural networks},
  author={You, Jiaxuan and Gomes-Selman, Jonathan M and Ying, Rex and Leskovec, Jure},
  booktitle={Proceedings of the AAAI conference on artificial intelligence},
  volume={35},
  number={12},
  pages={10737--10745},
  year={2021}
}

@article{cotta2021reconstruction,
  title={Reconstruction for powerful graph representations},
  author={Cotta, Leonardo and Morris, Christopher and Ribeiro, Bruno},
  journal={Advances in Neural Information Processing Systems},
  volume={34},
  pages={1713--1726},
  year={2021}
}

@article{zhang2021nested,
  title={Nested graph neural networks},
  author={Zhang, Muhan and Li, Pan},
  journal={Advances in Neural Information Processing Systems},
  volume={34},
  pages={15734--15747},
  year={2021}
}

@inproceedings{qianordered2022,
  title={Ordered Subgraph Aggregation Networks},
  author={Qian, Chendi and Rattan, Gaurav and Geerts, Floris and Niepert, Mathias and Morris, Christopher},
  booktitle={Advances in Neural Information Processing Systems},
  year={2022}
}

@article{zhou2023distance,
  title={Distance-Restricted Folklore Weisfeiler-Leman GNNs with Provable Cycle Counting Power},
  author={Zhou, Junru and Feng, Jiarui and Wang, Xiyuan and Zhang, Muhan},
  journal={arXiv preprint arXiv:2309.04941},
  year={2023}
}

@article{zhang2023complete,
  title={A Complete Expressiveness Hierarchy for Subgraph GNNs via Subgraph Weisfeiler-Lehman Tests},
  author={Zhang, Bohang and Feng, Guhao and Du, Yiheng and He, Di and Wang, Liwei},
  journal={arXiv preprint arXiv:2302.07090},
  year={2023}
}

@InProceedings{pmlr-v202-zhou23n,
  title = 	 {From Relational Pooling to Subgraph {GNN}s: A Universal Framework for More Expressive Graph Neural Networks},
  author =       {Zhou, Cai and Wang, Xiyuan and Zhang, Muhan},
  booktitle = 	 {Proceedings of the 40th International Conference on Machine Learning},
  pages = 	 {42742--42768},
  year = 	 {2023},
  volume = 	 {202},
  series = 	 {Proceedings of Machine Learning Research},
  publisher =    {PMLR}
}

@article{Kim2023pure,
  title={Pure Transformers are Powerful Graph Learners},
  author={Jinwoo Kim and Tien Dat Nguyen and Seonwoo Min and Sungjun Cho and Moontae Lee and Honglak Lee and Seunghoon Hong},
  journal={ArXiv},
  year={2022},
  volume={abs/2207.02505}
}

@InProceedings{pmlr-v238-zhou24a,
  title = 	 {On the Theoretical Expressive Power and the Design Space of Higher-Order Graph Transformers},
  author =       {Zhou, Cai and Yu, Rose and Wang, Yusu},
  booktitle = 	 {Proceedings of The 27th International Conference on Artificial Intelligence and Statistics},
  pages = 	 {2179--2187},
  year = 	 {2024},
  volume = 	 {238},
  series = 	 {Proceedings of Machine Learning Research},
  publisher =    {PMLR},
}

@article{bodnar2021weisfeiler,
  title={Weisfeiler and lehman go cellular: Cw networks},
  author={Bodnar, Cristian and Frasca, Fabrizio and Otter, Nina and Wang, Yuguang and Lio, Pietro and Montufar, Guido F and Bronstein, Michael},
  journal={Advances in Neural Information Processing Systems},
  volume={34},
  pages={2625--2640},
  year={2021}
}

@inproceedings{frascaunderstanding2022,
  title={Understanding and Extending Subgraph GNNs by Rethinking Their Symmetries},
  author={Frasca, Fabrizio and Bevilacqua, Beatrice and Bronstein, Michael M and Maron, Haggai},
  booktitle={Advances in Neural Information Processing Systems},
   year={2022}
}

@article{maron2018invariant,
  title={Invariant and equivariant graph networks},
  author={Maron, Haggai and Ben-Hamu, Heli and Shamir, Nadav and Lipman, Yaron},
  journal={arXiv preprint arXiv:1812.09902},
  year={2018}
}

@article{maron2019provably,
  title={Provably powerful graph networks},
  author={Maron, Haggai and Ben-Hamu, Heli and Serviansky, Hadar and Lipman, Yaron},
  journal={Advances in neural information processing systems},
  volume={32},
  year={2019}
}

@article{feng2023towards,
  title={Towards Arbitrarily Expressive GNNs in  {O}(n\^{} 2)  Space by Rethinking Folklore Weisfeiler-Lehman},
  author={Feng, Jiarui and Kong, Lecheng and Liu, Hao and Tao, Dacheng and Li, Fuhai and Zhang, Muhan and Chen, Yixin},
  journal={arXiv preprint arXiv:2306.03266},
  year={2023}
}

@article{chen2020can,
  title={Can graph neural networks count substructures?},
  author={Chen, Zhengdao and Chen, Lei and Villar, Soledad and Bruna, Joan},
  journal={Advances in neural information processing systems},
  volume={33},
  pages={10383--10395},
  year={2020}
}

@misc{li2020distance,
      title={Distance Encoding: Design Provably More Powerful Neural Networks for Graph Representation Learning}, 
      author={Pan Li and Yanbang Wang and Hongwei Wang and Jure Leskovec},
      year={2020},
      eprint={2009.00142},
      archivePrefix={arXiv},
      primaryClass={cs.LG}
}

@inbook{Rattan_2023,
   title={Weisfeiler-Leman and Graph Spectra},
   ISBN={9781611977554},
   url={http://dx.doi.org/10.1137/1.9781611977554.ch87},
   DOI={10.1137/1.9781611977554.ch87},
   booktitle={Proceedings of the 2023 Annual ACM-SIAM Symposium on Discrete Algorithms (SODA)},
   publisher={Society for Industrial and Applied Mathematics},
   author={Rattan, Gaurav and Seppelt, Tim},
   year={2023},
   month=jan, pages={2268–2285} 
}

@book{cvetkovic1997eigenspaces,
  title={Eigenspaces of graphs},
  author={Cvetkovi{\'c}, Drago{\v{s}} M and Rowlinson, Peter and Simic, Slobodan},
  number={66},
  year={1997},
  publisher={Cambridge University Press}
}

@article{FURER20102373,
title = {On the power of combinatorial and spectral invariants},
journal = {Linear Algebra and its Applications},
volume = {432},
number = {9},
pages = {2373-2380},
year = {2010},
note = {Special Issue devoted to Selected Papers presented at the Workshop on Spectral Graph Theory with Applications on Computer Science, Combinatorial Optimization and Chemistry (Rio de Janeiro, 2008)},
issn = {0024-3795},
doi = {https://doi.org/10.1016/j.laa.2009.07.019},
url = {https://www.sciencedirect.com/science/article/pii/S0024379509003620},
author = {Martin Fürer}
}

@misc{barcelo2021graph,
      title={Graph Neural Networks with Local Graph Parameters}, 
      author={Pablo Barceló and Floris Geerts and Juan Reutter and Maksimilian Ryschkov},
      year={2021},
      eprint={2106.06707},
      archivePrefix={arXiv},
      primaryClass={cs.LG}
}

@misc{kong2023maggnn,
      title={MAG-GNN: Reinforcement Learning Boosted Graph Neural Network}, 
      author={Lecheng Kong and Jiarui Feng and Hao Liu and Dacheng Tao and Yixin Chen and Muhan Zhang},
      year={2023},
      eprint={2310.19142},
      archivePrefix={arXiv},
      primaryClass={cs.LG}
}

@misc{bevilacqua2024efficient,
      title={Efficient Subgraph GNNs by Learning Effective Selection Policies}, 
      author={Beatrice Bevilacqua and Moshe Eliasof and Eli Meirom and Bruno Ribeiro and Haggai Maron},
      year={2024},
      eprint={2310.20082},
      archivePrefix={arXiv},
      primaryClass={cs.LG}
}

@article{bouritsas2022improving,
  title={Improving graph neural network expressivity via subgraph isomorphism counting},
  author={Bouritsas, Giorgos and Frasca, Fabrizio and Zafeiriou, Stefanos and Bronstein, Michael M},
  journal={IEEE Transactions on Pattern Analysis and Machine Intelligence},
  volume={45},
  number={1},
  pages={657--668},
  year={2022},
  publisher={IEEE}
}

@article{kipf2016semi,
  title={Semi-supervised classification with graph convolutional networks},
  author={Kipf, Thomas N and Welling, Max},
  journal={arXiv preprint arXiv:1609.02907},
  year={2016}
}

@inproceedings{zaheer2017deep,
author = {Zaheer, Manzil and Kottur, Satwik and Ravanbhakhsh, Siamak and P\'{o}czos, Barnab\'{a}s and Salakhutdinov, Ruslan and Smola, Alexander J},
title = {Deep Sets},
year = {2017},
isbn = {9781510860964},
publisher = {Curran Associates Inc.},
address = {Red Hook, NY, USA},
booktitle = {Proceedings of the 31st International Conference on Neural Information Processing Systems},
pages = {3394–3404},
numpages = {11},
location = {Long Beach, California, USA},
series = {NIPS'17}
}

@article{bronstein2021geometric,
  title={Geometric deep learning: Grids, groups, graphs, geodesics, and gauges},
  author={Bronstein, Michael M and Bruna, Joan and Cohen, Taco and Veli{\v{c}}kovi{\'c}, Petar},
  journal={arXiv preprint arXiv:2104.13478},
  year={2021}
}

@article{villar2021scalars,
  title={Scalars are universal: Equivariant machine learning, structured like classical physics},
  author={Villar, Soledad and Hogg, David W and Storey-Fisher, Kate and Yao, Weichi and Blum-Smith, Ben},
  journal={Advances in Neural Information Processing Systems},
  volume={34},
  pages={28848--28863},
  year={2021}
}

@article{gasteiger2020directional,
  title={Directional message passing for molecular graphs},
  author={Gasteiger, Johannes and Gro{\ss}, Janek and G{\"u}nnemann, Stephan},
  journal={arXiv preprint arXiv:2003.03123},
  year={2020}
}

@article{li2024distance,
  title={Is distance matrix enough for geometric deep learning?},
  author={Li, Zian and Wang, Xiyuan and Huang, Yinan and Zhang, Muhan},
  journal={Advances in Neural Information Processing Systems},
  volume={36},
  year={2024}
}

@article{gasteiger2021gemnet,
  title={Gemnet: Universal directional graph neural networks for molecules},
  author={Gasteiger, Johannes and Becker, Florian and G{\"u}nnemann, Stephan},
  journal={Advances in Neural Information Processing Systems},
  volume={34},
  pages={6790--6802},
  year={2021}
}

@article{yu2015useful,
  title={A useful variant of the Davis--Kahan theorem for statisticians},
  author={Yu, Yi and Wang, Tengyao and Samworth, Richard J},
  journal={Biometrika},
  volume={102},
  number={2},
  pages={315--323},
  year={2015},
  publisher={Oxford University Press}
}

@inproceedings{deng2021vector,
  title={Vector neurons: A general framework for so (3)-equivariant networks},
  author={Deng, Congyue and Litany, Or and Duan, Yueqi and Poulenard, Adrien and Tagliasacchi, Andrea and Guibas, Leonidas J},
  booktitle={Proceedings of the IEEE/CVF International Conference on Computer Vision},
  pages={12200--12209},
  year={2021}
}

@inproceedings{satorras2021n,
  title={E (n) equivariant graph neural networks},
  author={Satorras, V{\i}ctor Garcia and Hoogeboom, Emiel and Welling, Max},
  booktitle={International conference on machine learning},
  pages={9323--9332},
  year={2021},
  organization={PMLR}
}

@article{thomas2018tensor,
  title={Tensor field networks: Rotation-and translation-equivariant neural networks for 3d point clouds},
  author={Thomas, Nathaniel and Smidt, Tess and Kearnes, Steven and Yang, Lusann and Li, Li and Kohlhoff, Kai and Riley, Patrick},
  journal={arXiv preprint arXiv:1802.08219},
  year={2018}
}

@inproceedings{bogatskiy2020lorentz,
  title={Lorentz group equivariant neural network for particle physics},
  author={Bogatskiy, Alexander and Anderson, Brandon and Offermann, Jan and Roussi, Marwah and Miller, David and Kondor, Risi},
  booktitle={International Conference on Machine Learning},
  pages={992--1002},
  year={2020},
  organization={PMLR}
}

@article{cohen2018spherical,
  title={Spherical cnns},
  author={Cohen, Taco S and Geiger, Mario and K{\"o}hler, Jonas and Welling, Max},
  journal={arXiv preprint arXiv:1801.10130},
  year={2018}
}

@article{batzner20223,
  title={E (3)-equivariant graph neural networks for data-efficient and accurate interatomic potentials},
  author={Batzner, Simon and Musaelian, Albert and Sun, Lixin and Geiger, Mario and Mailoa, Jonathan P and Kornbluth, Mordechai and Molinari, Nicola and Smidt, Tess E and Kozinsky, Boris},
  journal={Nature communications},
  volume={13},
  number={1},
  pages={2453},
  year={2022},
  publisher={Nature Publishing Group UK London}
}

@inproceedings{schutt2021equivariant,
  title={Equivariant message passing for the prediction of tensorial properties and molecular spectra},
  author={Sch{\"u}tt, Kristof and Unke, Oliver and Gastegger, Michael},
  booktitle={International Conference on Machine Learning},
  pages={9377--9388},
  year={2021},
  organization={PMLR}
}

@article{fuchs2020se,
  title={Se (3)-transformers: 3d roto-translation equivariant attention networks},
  author={Fuchs, Fabian and Worrall, Daniel and Fischer, Volker and Welling, Max},
  journal={Advances in neural information processing systems},
  volume={33},
  pages={1970--1981},
  year={2020}
}

@inproceedings{finkelshtein2022simple,
  title={A simple and universal rotation equivariant point-cloud network},
  author={Finkelshtein, Ben and Baskin, Chaim and Maron, Haggai and Dym, Nadav},
  booktitle={Topological, Algebraic and Geometric Learning Workshops 2022},
  pages={107--115},
  year={2022},
  organization={PMLR}
}

@article{ruhe2024clifford,
  title={Clifford group equivariant neural networks},
  author={Ruhe, David and Brandstetter, Johannes and Forr{\'e}, Patrick},
  journal={Advances in Neural Information Processing Systems},
  volume={36},
  year={2024}
}

@article{dym2020universality,
  title={On the universality of rotation equivariant point cloud networks},
  author={Dym, Nadav and Maron, Haggai},
  journal={arXiv preprint arXiv:2010.02449},
  year={2020}
}

@article{wu2020comprehensive,
  title={A comprehensive survey on graph neural networks},
  author={Wu, Zonghan and Pan, Shirui and Chen, Fengwen and Long, Guodong and Zhang, Chengqi and Philip, S Yu},
  journal={IEEE transactions on neural networks and learning systems},
  volume={32},
  number={1},
  pages={4--24},
  year={2020},
  publisher={IEEE}
}

@article{zhou2020graph,
  title={Graph neural networks: A review of methods and applications},
  author={Zhou, Jie and Cui, Ganqu and Hu, Shengding and Zhang, Zhengyan and Yang, Cheng and Liu, Zhiyuan and Wang, Lifeng and Li, Changcheng and Sun, Maosong},
  journal={AI open},
  volume={1},
  pages={57--81},
  year={2020},
  publisher={Elsevier}
}

@article{reiser2022graph,
  title={Graph neural networks for materials science and chemistry},
  author={Reiser, Patrick and Neubert, Marlen and Eberhard, Andr{\'e} and Torresi, Luca and Zhou, Chen and Shao, Chen and Metni, Houssam and van Hoesel, Clint and Schopmans, Henrik and Sommer, Timo and others},
  journal={Communications Materials},
  volume={3},
  number={1},
  pages={93},
  year={2022},
  publisher={Nature Publishing Group UK London}
}

@techreport{deshpande2002automated,
  title={Automated approaches for classifying structures},
  author={Deshpande, Mukund and Kuramochi, Michihiro and Karypis, George},
  year={2002},
  institution={MINNESOTA UNIV MINNEAPOLIS DEPT OF COMPUTER SCIENCE}
}

@inproceedings{jin2018junction,
  title={Junction tree variational autoencoder for molecular graph generation},
  author={Jin, Wengong and Barzilay, Regina and Jaakkola, Tommi},
  booktitle={International conference on machine learning},
  pages={2323--2332},
  year={2018},
  organization={PMLR}
}

@article{stokes2020deep,
  title={A deep learning approach to antibiotic discovery},
  author={Stokes, Jonathan M and Yang, Kevin and Swanson, Kyle and Jin, Wengong and Cubillos-Ruiz, Andres and Donghia, Nina M and MacNair, Craig R and French, Shawn and Carfrae, Lindsey A and Bloom-Ackermann, Zohar and others},
  journal={Cell},
  volume={180},
  number={4},
  pages={688--702},
  year={2020},
  publisher={Elsevier}
}

@inproceedings{ying2018graph,
  title={Graph convolutional neural networks for web-scale recommender systems},
  author={Ying, Rex and He, Ruining and Chen, Kaifeng and Eksombatchai, Pong and Hamilton, William L and Leskovec, Jure},
  booktitle={Proceedings of the 24th ACM SIGKDD international conference on knowledge discovery \& data mining},
  pages={974--983},
  year={2018}
}

@article{zitnik2017predicting,
  title={Predicting multicellular function through multi-layer tissue networks},
  author={Zitnik, Marinka and Leskovec, Jure},
  journal={Bioinformatics},
  volume={33},
  number={14},
  pages={i190--i198},
  year={2017},
  publisher={Oxford University Press}
}

@article{zitnik2018modeling,
  title={Modeling polypharmacy side effects with graph convolutional networks},
  author={Zitnik, Marinka and Agrawal, Monica and Leskovec, Jure},
  journal={Bioinformatics},
  volume={34},
  number={13},
  pages={i457--i466},
  year={2018},
  publisher={Oxford University Press}
}

@inproceedings{lopera2021survey,
  title={A survey of graph neural networks for electronic design automation},
  author={Lopera, Daniela S{\'a}nchez and Servadei, Lorenzo and Kiprit, Gamze Naz and Hazra, Souvik and Wille, Robert and Ecker, Wolfgang},
  booktitle={2021 ACM/IEEE 3rd Workshop on Machine Learning for CAD (MLCAD)},
  pages={1--6},
  year={2021},
  organization={IEEE}
}

@article{zhang2021labeling,
  title={Labeling trick: A theory of using graph neural networks for multi-node representation learning},
  author={Zhang, Muhan and Li, Pan and Xia, Yinglong and Wang, Kai and Jin, Long},
  journal={Advances in Neural Information Processing Systems},
  volume={34},
  pages={9061--9073},
  year={2021}
}

@article{alon2020bottleneck,
  title={On the bottleneck of graph neural networks and its practical implications},
  author={Alon, Uri and Yahav, Eran},
  journal={arXiv preprint arXiv:2006.05205},
  year={2020}
}

@article{dwivedi2022long,
  title={Long range graph benchmark},
  author={Dwivedi, Vijay Prakash and Ramp{\'a}{\v{s}}ek, Ladislav and Galkin, Michael and Parviz, Ali and Wolf, Guy and Luu, Anh Tuan and Beaini, Dominique},
  journal={Advances in Neural Information Processing Systems},
  volume={35},
  pages={22326--22340},
  year={2022}
}

@article{von2007tutorial,
  title={A tutorial on spectral clustering},
  author={Von Luxburg, Ulrike},
  journal={Statistics and computing},
  volume={17},
  pages={395--416},
  year={2007},
  publisher={Springer}
}

@article{ying2021transformers,
  title={Do transformers really perform badly for graph representation?},
  author={Ying, Chengxuan and Cai, Tianle and Luo, Shengjie and Zheng, Shuxin and Ke, Guolin and He, Di and Shen, Yanming and Liu, Tie-Yan},
  journal={Advances in neural information processing systems},
  volume={34},
  pages={28877--28888},
  year={2021}
}

@inproceedings{chen2022structure,
  title={Structure-aware transformer for graph representation learning},
  author={Chen, Dexiong and O’Bray, Leslie and Borgwardt, Karsten},
  booktitle={International Conference on Machine Learning},
  pages={3469--3489},
  year={2022},
  organization={PMLR}
}

@inproceedings{specformer,
  title={Specformer: Spectral Graph Neural Networks Meet Transformers},
  author={Deyu Bo and Chuan Shi and Lele Wang and Renjie Liao},
  booktitle={The Eleventh International Conference on Learning Representations },
  year={2023}
}

@inproceedings{huang2023boosting,
  title={Boosting the Cycle Counting Power of Graph Neural Networks with {I}$^2$-{GNN}s},
  author={Yinan Huang and Xingang Peng and Jianzhu Ma and Muhan Zhang},
  booktitle={The Eleventh International Conference on Learning Representations },
  year={2023}
}

@inproceedings{Khop,
  title={How Powerful are K-hop Message Passing Graph Neural Networks},
  author={Jiarui Feng and Yixin Chen and Fuhai Li and Anindya Sarkar and Muhan Zhang},
  booktitle={Advances in Neural Information Processing Systems},
  year={2022}
}

@article{QM9,
  title={Quantum chemistry structures and properties of 134 kilo molecules},
  author={Raghunathan Ramakrishnan and Pavlo O. Dral and Matthias Rupp and O. Anatole von Lilienfeld},
  journal={Scientific Data},
  year={2014},
  volume={1}
}

@article{Zinc,
  title={Benchmarking Graph Neural Networks},
  author={Vijay Prakash Dwivedi and Chaitanya K. Joshi and Thomas Laurent and Yoshua Bengio and Xavier Bresson},
  journal={ArXiv},
  year={2020},
  volume={abs/2003.00982}
}

@article{DTNN,
  title={MoleculeNet: a benchmark for molecular machine learning† †Electronic supplementary information (ESI) available. See DOI: 10.1039/c7sc02664a},
  author={Zhenqin Wu and Bharath Ramsundar and Evan N. Feinberg and Joseph Gomes and Caleb Geniesse and Aneesh S. Pappu and Karl Leswing and Vijay S. Pande},
  journal={Chemical Science},
  year={2017},
  volume={9},
  pages={513 - 530}
}

@inproceedings{HodgeRandomWalk,
 author = {Zhou, Cai and Wang, Xiyuan and Zhang, Muhan},
 booktitle = {Advances in Neural Information Processing Systems},
 pages = {16172--16206},
 title = {Facilitating Graph Neural Networks with Random Walk on Simplicial Complexes},
 volume = {36},
 year = {2023}
}

@article{PNA,
  title={Principal Neighbourhood Aggregation for Graph Nets},
  author={Gabriele Corso and Luca Cavalleri and D. Beaini and Pietro Lio’ and Petar Velickovic},
  journal={ArXiv},
  year={2020},
  volume={abs/2004.05718}
}

@article{GNNAK,
  title={From Stars to Subgraphs: Uplifting Any GNN with Local Structure Awareness},
  author={Lingxiao Zhao and Wei Jin and Leman Akoglu and Neil Shah},
  journal={ArXiv},
  year={2021},
  volume={abs/2110.03753}
}

@inproceedings{Graphormer,
 author = {Ying, Chengxuan and Cai, Tianle and Luo, Shengjie and Zheng, Shuxin and Ke, Guolin and He, Di and Shen, Yanming and Liu, Tie-Yan},
 booktitle = {Advances in Neural Information Processing Systems},
 pages = {28877--28888},
 title = {Do Transformers Really Perform Badly for Graph Representation?},
 volume = {34},
 year = {2021}
}

@article{PretrainGINE,
  title={Strategies for Pre-training Graph Neural Networks},
  author={Weihua Hu and Bowen Liu and Joseph Gomes and Marinka Zitnik and Percy Liang and Vijay S. Pande and Jure Leskovec},
  journal={arXiv: Learning},
  year={2019}
}

@article{ResidualGatedGCN,
  title={Residual Gated Graph ConvNets},
  author={Xavier Bresson and Thomas Laurent},
  journal={ArXiv},
  year={2017},
  volume={abs/1711.07553}
}

@article{wang2023pytorch,
  title={PyTorch Geometric High Order: A Unified Library for High Order Graph Neural Network},
  author={Wang, Xiyuan and Zhang, Muhan},
  journal={arXiv preprint arXiv:2311.16670},
  year={2023}
}

\newpage
\appendix
\onecolumn

\section{Proof of Proposition~\ref{prop:stability_bound}}\label{app:proof-stable}

Before proving Proposition~\ref{prop:stability_bound}, we present some useful lemmas. We quote these lemmas directly from~\citep{huang2023stability}. 

\begin{lemma}[Davis-Kahan theorem, Proposition A.1 of~\citep{huang2023stability}, see also~\citep{yu2015useful}]\label{lemma:davis-kahan}
    Let $\bm{A}, \bm{A}'$ be $n\times n$ real symmetric matrices. Let $\lambda_1\leqslant \cdots\leqslant \lambda_n$ be eigenvalues of $\bm{A}$ sorted in increasing order (possibly with repeats). Let the columns of $\bm{V}, \bm{V}'\in O(n)$ contain mutually orthogonal normalized eigenvectors of $\bm{A}, \bm{A}'$ respectively, sorted in increasing order of their corresponding eigenvalues. Let $\mathcal{J}=\{s, s+1,\ldots, t\}\subseteq\{1,\ldots, n\}$ be a contiguous interval of indices, and $[\bm{V}]_\mathcal{J}, [\bm{V}']_\mathcal{J}$ be matrices of shape $n\times |\mathcal{J}|$ whose columns are the $s$-th, $(s+1)$-th, $\ldots$, $t$-th column of $\bm{V}$ and $\bm{V}'$, respectively. Then
    \begin{align}
        \min_{\bm{Q}\in O(|\mathcal{J}|)} \|[\bm{V}]_\mathcal{J}-[\bm{V}']_\mathcal{J}\bm{Q}\|_\text{F} \leqslant \frac{\sqrt{8}\min\left\{\sqrt{|\mathcal{J}|}\|\bm{A}-\bm{A}'\|_2, \|\bm{A}-\bm{A}'\|_\text{F}\right\}}{\min\{\lambda_s-\lambda_{s-1}, \lambda_{t+1}-\lambda_t\}}.
    \end{align}
    For convenience, we define $\lambda_0=-\infty$ and $\lambda_{n+1}=+\infty$.
\end{lemma}

\begin{lemma}[Weyl's inequality, Proposition A.2 of~\citep{huang2023stability}]\label{lemma:weyl}
    Given a real symmetric matrix $\bm{A}$, let $\lambda_i(\bm{A})$ be its $i$-th smallest eigenvalue. For any two real symmetric matrices $\bm{A}, \bm{A}'$ of shape $n\times n$, $|\lambda_i(\bm{A})-\lambda_i(\bm{A}')|\leqslant \|\bm{A}-\bm{A}'\|_2$ holds for all $i=1,\ldots, n$.
\end{lemma}

\begin{lemma}[Lemma A.1 of~\citep{huang2023stability}]\label{lemma:product}
    Assume $\bm{A}_1\bm{A}_2\cdots \bm{A}_p$ is a valid matrix multiplication. Then
    \begin{align}
        \left\|\prod_{k=1}^p \bm{A}_k\right\|_\text{F}\leqslant \left(\prod_{k=1}^{\ell-1}\|\bm{A}_k\|_2\right)\|\bm{A}_\ell\|_\text{F}\left(\prod_{k=\ell+1}^p\|\bm{A}_k^T\|_2\right).
    \end{align}
\end{lemma}

Now we can present the proof of Proposition~\ref{prop:stability_bound}.

\begin{proof}
    We will prove the uniform result that for any two graphs $G, G'\in\mathcal{G}$ with Laplacians $\bm{L}, \bm{L}'$ respectively, and for any $\bm{P}\in S_n$, there exists a value of $\delta$ such that
    \begin{align}
        \|Z(G)-\bm{P}Z(G')\|_\text{F} &\leqslant nJ_\psi J_\phi\big[(\sqrt{n}+2nJ_\rho J_f)\|\bm{L}-\bm{PL}'\bm{P}^T\|_2
        +4\sqrt[4]{2}J_f\sqrt{J_\rho}n\|\bm{L}-\bm{PL}'\bm{P}^T\|_\text{F}^{1/2}\big].
    \end{align} 
    We may first rewrite equation~\eqref{eq:smooth1} as
    \begin{align}
        Z(G) = \psi\left(\sum_{i=1}^n\phi\left(\mathrm{concat}\left[ \tilde{\lambda}_i\mathbf{1}_{n}, f(\tilde{\bm{V}}_i^\text{smooth})\right]\right)\right),\label{eq:refined_zg}
    \end{align}
    in which $\tilde{\lambda}_i$ is the $i$-th smallest eigenvalue of $\bm{L}$ (\emph{including repeats} when counting orders), and
    \begin{align}\label{eq:ineedit}
        \tilde{\bm{V}}_i^{\text{smooth}}=\mathrm{concat}\left[\bm{v}_1\rho(|\tilde{\lambda}_1-\tilde{\lambda}_i|),\bm{v}_2\rho(|\tilde{\lambda}_2-\tilde{\lambda}_i|),\ldots, \bm{v}_n\rho(|\tilde{\lambda}_n-\tilde{\lambda}_i|)\right],
    \end{align}
    where column vectors $\bm{v}_1,\bm{v}_2,\ldots, \bm{v}_n\in\mathbb{R}^{n\times 1}$ are mutually orthogonal normalized eigenvectors corresponding to eigenvalues $\tilde{\lambda}_1,\tilde{\lambda}_2,\ldots,\tilde{\lambda}_n$ respectively. With equation~\eqref{eq:refined_zg}, we have completely removed the dependency on eigenspace dimensions in $Z(G)$. We then have
    \begin{align}
        \|Z(G)-\bm{P}Z(G')\|_\text{F} &= \left\|\psi\left(\sum_{i=1}^n\phi\left(\mathrm{concat}\left[ \tilde{\lambda}_i\mathbf{1}_{n}, f(\tilde{\bm{V}}_i^\text{smooth})\right]\right)\right)\right.
        -\left.\bm{P}\psi\left(\sum_{i=1}^n\phi\left(\mathrm{concat}\left[ \tilde{\lambda}'_i\mathbf{1}_{n}, f(\tilde{\bm{V}}_i'^{\text{smooth}})\right]\right)\right)\right\|_\text{F}\\
        &=\left\|\psi\left(\sum_{i=1}^n\phi\left(\mathrm{concat}\left[ \tilde{\lambda}_i\mathbf{1}_{n}, f(\tilde{\bm{V}}_i^\text{smooth})\right]\right)\right)\right.
        -\left.\psi\left(\sum_{i=1}^n\phi\left(\mathrm{concat}\left[ \tilde{\lambda}'_i\mathbf{1}_{n}, f(\bm{P}\tilde{\bm{V}}_i'^{\text{smooth}})\right]\right)\right)\right\|_\text{F}\label{eq:_first_step}\\
        &\leqslant J_\psi\left\|\sum_{i=1}^n\phi\left(\mathrm{concat}\left[ \tilde{\lambda}_i\mathbf{1}_{n}, f(\tilde{\bm{V}}_i^\text{smooth})\right]\right)\right.
        -\left.\sum_{i=1}^n\phi\left(\mathrm{concat}\left[ \tilde{\lambda}'_i\mathbf{1}_{n}, f(\bm{P}\tilde{\bm{V}}_i'^{\text{smooth}})\right]\right)\right\|_\text{F}\label{eq:_second_step}\\
        & \leqslant J_\psi\sum_{i=1}^n\left\|\phi\left(\mathrm{concat}\left[ \tilde{\lambda}_i\mathbf{1}_{n}, f(\tilde{\bm{V}}_i^\text{smooth})\right]\right)\right.
        -\left.\phi\left(\mathrm{concat}\left[ \tilde{\lambda}'_i\mathbf{1}_{n}, f(\bm{P}\tilde{\bm{V}}_i'^{\text{smooth}})\right]\right)\right\|_\text{F}\label{eq:_third_step}\\
        &\leqslant J_\psi J_\phi\sum_{i=1}^n\left\|\mathrm{concat}\left[ \left(\tilde{\lambda}_i-\tilde{\lambda}'_i\right)\mathbf{1}_{n}, f(\tilde{\bm{V}}_i^\text{smooth})-f(\bm{P}\tilde{\bm{V}}_i'^{\text{smooth}})\right]\right\|_\text{F}\label{eq:_fourth_step}\\
        &\leqslant J_\psi J_\phi\sum_{i=1}^n \left[\sqrt{n}\left|\tilde{\lambda}_i-\tilde{\lambda}'_i\right|+\|f(\tilde{\bm{V}}_i^\text{smooth})-f(\bm{P}\tilde{\bm{V}}_i'^{\text{smooth}})\|\right].\label{eq:_fifth_step}
    \end{align}
    The equality on~\eqref{eq:_first_step} is due to the fact that $\psi$ and $\phi$ operate row-wise on the $n$ rows of their arguments, and that $f$ is permutation equivariant. \eqref{eq:_second_step} and \eqref{eq:_fourth_step} stem from the Lipschitz continuities of $\psi$ and $\phi$, respectively. \eqref{eq:_third_step} is due to triangular inequality. Now it suffices to bound the two terms in~\eqref{eq:_fifth_step}.

    For the first term, we invoke Lemma~\ref{lemma:weyl} to get
    \begin{align}
        \sum_{i=1}^n\left|\tilde{\lambda}_i-\tilde{\lambda}'_i\right|\leqslant \sum_{i=1}^n\|\bm{L}-\bm{PL}'\bm{P}^T\|_2=n\|\bm{L}-\bm{PL}'\bm{P}^T\|_2,\quad \forall \bm{P}\in S_n.
    \end{align}
    This is because for any $\bm{P}\in S_n$, $\bm{PL}'\bm{P}^T$ has the same sequence of eigenvalues as $\bm{L}'$, namely $\tilde{\lambda}'_1,\tilde{\lambda}'_2,\ldots, \tilde{\lambda}'_n$.

    For the second term, we have
    \begin{align}
        \|f(\tilde{\bm{V}}_i^\text{smooth})-f(\bm{P}\tilde{\bm{V}}_i'^{\text{smooth}})\|&\leqslant J_f\min_{\bm{Q}_i\in O(n)}\|\tilde{\bm{V}}_i^\text{smooth}-\bm{P}\tilde{\bm{V}}_i'^{\text{smooth}}\bm{Q}_i\|_\text{F}\label{eq:__first__step}\\
        \notag&=J_f\min_{\bm{Q}_i\in O(n)}\left\|\mathrm{concat}\left[\bm{v}_1\rho(|\tilde{\lambda}_1-\tilde{\lambda}_i|),\ldots, \bm{v}_n\rho(|\tilde{\lambda}_n-\tilde{\lambda}_i|)\right]\right.\\
        &\quad-\left.\mathrm{concat}\left[\bm{Pv}'_1\rho(|\tilde{\lambda}'_1-\tilde{\lambda}'_i|),\ldots, \bm{Pv}'_n\rho(|\tilde{\lambda}'_n-\tilde{\lambda}'_i|)\right]\bm{Q}_i\right\|_\text{F}\label{eq:__second__step}\\
        \notag&\leqslant J_f\min_{\bm{Q}_i\in O(n)}\left\{\left\|\mathrm{concat}\left[\bm{v}_1\rho(|\tilde{\lambda}_1-\tilde{\lambda}_i|),\ldots, \bm{v}_n\rho(|\tilde{\lambda}_n-\tilde{\lambda}_i|)\right]\right.\right.\\
        \notag&\quad-\left.\mathrm{concat}\left[\bm{v}_1\rho(|\tilde{\lambda}'_1-\tilde{\lambda}'_i|),\ldots, \bm{v}_n\rho(|\tilde{\lambda}'_n-\tilde{\lambda}'_i|)\right]\right\|_\text{F}\\
        \notag&\quad +\left\|\mathrm{concat}\left[\bm{v}_1\rho(|\tilde{\lambda}'_1-\tilde{\lambda}'_i|),\ldots, \bm{v}_n\rho(|\tilde{\lambda}'_n-\tilde{\lambda}'_i|)\right]\right.\\
        &\quad-\left.\left.\mathrm{concat}\left[\bm{Pv}'_1\rho(|\tilde{\lambda}'_1-\tilde{\lambda}'_i|),\ldots, \bm{Pv}'_n\rho(|\tilde{\lambda}'_n-\tilde{\lambda}'_i|)\right]\bm{Q}_i\right\|_\text{F}\right\}\label{eq:__third__step}\\
        \notag &=J_f\sqrt{\sum_{j=1}^n\left[\rho(|\tilde{\lambda}_j-\tilde{\lambda}_i|)-\rho(|\tilde{\lambda}'_j-\tilde{\lambda}'_i|)\right]^2}\\
        \notag &\quad+J_f\min_{\bm{Q}_i\in O(n)}\left\|\mathrm{concat}\left[\bm{v}_1\rho(|\tilde{\lambda}'_1-\tilde{\lambda}'_i|),\ldots, \bm{v}_n\rho(|\tilde{\lambda}'_n-\tilde{\lambda}'_i|)\right]\right.\\
        &\quad-\left.\mathrm{concat}\left[\bm{Pv}'_1\rho(|\tilde{\lambda}'_1-\tilde{\lambda}'_i|),\ldots, \bm{Pv}'_n\rho(|\tilde{\lambda}'_n-\tilde{\lambda}'_i|)\right]\bm{Q}_i\right\|_\text{F}.\label{eq:__fourth__step}
    \end{align}
    Here, \eqref{eq:__first__step} is due to our assumption on $f$, \eqref{eq:__second__step} follows from definitions of $\tilde{\bm{V}}^\text{smooth}_i$ and $\tilde{\bm{V}}'^{\text{smooth}}_i$, while \eqref{eq:__third__step} stems from triangular inequality. Now, for the first term of \eqref{eq:__fourth__step}, we have
    \begin{align}
        \sqrt{\sum_{j=1}^n\left[\rho(|\tilde{\lambda}_j-\tilde{\lambda}_i|)-\rho(|\tilde{\lambda}'_j-\tilde{\lambda}'_i|)\right]^2}&\leqslant \sum_{j=1}^n\left|\rho(|\tilde{\lambda}_j-\tilde{\lambda}_i|)-\rho(|\tilde{\lambda}'_j-\tilde{\lambda}'_i|)\right|\\
        &\leqslant J_\rho \sum_{j=1}^n\left||\tilde{\lambda}_j-\tilde{\lambda}_i|-|\tilde{\lambda}'_j-\tilde{\lambda}'_i|\right|\label{eq:___first___step}\\
        &\leqslant J_\rho \sum_{j=1}^n\left(|\tilde{\lambda}_i-\tilde{\lambda}'_i|+|\tilde{\lambda}_j-\tilde{\lambda}'_j|\right)\label{eq:___second___step}\\
        &\leqslant 2nJ_\rho \|\bm{L}-\bm{PL}'\bm{P}^T\|_2,\quad \forall \bm{P}\in S_n,\label{eq:___third___step}
    \end{align}
    where \eqref{eq:___first___step} is by Lipschitz continuity of $\rho$, \eqref{eq:___second___step} makes use of the fact that either $\tilde{\lambda}_i\geqslant \tilde{\lambda}_j$ and $\tilde{\lambda}'_i\geqslant \tilde{\lambda}'_j$, or $\tilde{\lambda}_i\leqslant \tilde{\lambda}_j$ and $\tilde{\lambda}'_i\leqslant \tilde{\lambda}'_j$. The final step~\eqref{eq:___third___step} stems from Lemma~\ref{lemma:weyl}.

    To bound the second term of \eqref{eq:__fourth__step}, we first split the eigenvalues $\tilde{\lambda}'_1,\tilde{\lambda}'_2,\ldots, \tilde{\lambda}'_n$ into groups, namely $\mathcal{J}_1 = \{\tilde{\lambda}'_{J_0+1},\ldots, \tilde{\lambda}'_{J_1}\}$, $\mathcal{J}_2 = \{\tilde{\lambda}'_{J_1+1},\ldots, \tilde{\lambda}'_{J_2}\}$, $\ldots$, $\mathcal{J}_L=\{\tilde{\lambda}'_{J_{L-1}+1},\ldots, \tilde{\lambda}'_{J_L}\}$, with $J_0=0$ and $J_L=n$. We ask that $\tilde{\lambda}'_{k+1}-\tilde{\lambda}'_k>\delta$ for all $k=J_0,J_1,\ldots, J_L$, and $\tilde{\lambda}'_{k+1}-\tilde{\lambda}'_k\leqslant\delta$ for all other $k$. We also denote by $\mathcal{J}(\tilde{\lambda}'_i)$ the group where $\tilde{\lambda}'_i$ belong.
    
    The consequence of such splitting is that for any eigenvalue $\tilde{\lambda}'_i$ of $\bm{L}'$, all $\tilde{\lambda}'_j$ satisfying $\rho(|\tilde{\lambda}'_j-\tilde{\lambda}'_i|)\ne 0$ belong to $\mathcal{J}(\tilde{\lambda}'_i)$. Therefore, we actually have
    \begin{align}
        \notag&\qquad\min_{\bm{Q}_i\in O(n)}\left\|\mathrm{concat}\left[\bm{v}_j\rho(|\tilde{\lambda}'_j-\tilde{\lambda}'_i|)\right]_{j=1}^n-\mathrm{concat}\left[\bm{Pv}'_j\rho(|\tilde{\lambda}'_j-\tilde{\lambda}'_i|)\right]_{j=1}^n\bm{Q}_i\right\|_\text{F}\\
        &=\min_{\bm{Q}_i\in O(|\mathcal{J}(\tilde{\lambda}'_i)|)}\left\|\mathrm{concat}\left[\bm{v}_j\rho(|\tilde{\lambda}'_j-\tilde{\lambda}'_i|)\right]_{\tilde{\lambda}'_j\in\mathcal{J}(\tilde{\lambda}'_i)}-\mathrm{concat}\left[\bm{Pv}'_j\rho(|\tilde{\lambda}'_j-\tilde{\lambda}'_i|)\right]_{\tilde{\lambda}'_j\in\mathcal{J}(\tilde{\lambda}'_i)}\bm{Q}_i\right\|_\text{F}.
    \end{align}
    Now, for any $\bm{Q}_i\in O(|\mathcal{J}(\tilde{\lambda}'_i)|)$, we have
    \begin{align}
        \notag &\quad \left\|\mathrm{concat}\left[\bm{v}_j\rho(|\tilde{\lambda}'_j-\tilde{\lambda}'_i|)\right]_{\tilde{\lambda}'_j\in\mathcal{J}(\tilde{\lambda}'_i)}-\mathrm{concat}\left[\bm{Pv}'_j\rho(|\tilde{\lambda}'_j-\tilde{\lambda}'_i|)\right]_{\tilde{\lambda}'_j\in\mathcal{J}(\tilde{\lambda}'_i)}\bm{Q}_i\right\|_\text{F}\\
        &=\left\|\mathrm{concat}\left[\bm{v}_j\rho(|\tilde{\lambda}'_j-\tilde{\lambda}'_i|)-\sum_{k:\tilde{\lambda}'_k\in\mathcal{J}(\tilde{\lambda}'_i)}\bm{Pv}'_k\rho(|\tilde{\lambda}'_k-\tilde{\lambda}'_i|)(\bm{Q}_i)_{kj}\right]_{\tilde{\lambda}'_j\in\mathcal{J}(\tilde{\lambda}'_i)}\right\|_\text{F}\\
        &\notag \leqslant \left\|\mathrm{concat}\left[\sum_{k:\tilde{\lambda}'_k\in\mathcal{J}(\tilde{\lambda}'_i)}\bm{Pv}'_k\left[\rho(|\tilde{\lambda}'_j-\tilde{\lambda}'_i|)-\rho(|\tilde{\lambda}'_k-\tilde{\lambda}'_i|)\right](\bm{Q}_i)_{kj}\right]_{\tilde{\lambda}'_j\in\mathcal{J}(\tilde{\lambda}'_i)}\right\|_\text{F}\\
        &\quad +\left\|\mathrm{concat}\left[\rho(|\tilde{\lambda}'_j-\tilde{\lambda}'_i|)\left(\bm{v}_j-\sum_{k:\tilde{\lambda}'_k\in\mathcal{J}(\tilde{\lambda}'_i)}\bm{Pv}'_k(\bm{Q}_i)_{kj}\right)\right]_{\tilde{\lambda}'_j\in\mathcal{J}(\tilde{\lambda}'_i)}\right\|_\text{F}\\
        &\notag \leqslant \sum_{j:\tilde{\lambda}'_j\in\mathcal{J}(\tilde{\lambda}'_i)}\left\|\sum_{k:\tilde{\lambda}'_k\in\mathcal{J}(\tilde{\lambda}'_i)}\bm{Pv}'_k\left[\rho(|\tilde{\lambda}'_j-\tilde{\lambda}'_i|)-\rho(|\tilde{\lambda}'_k-\tilde{\lambda}'_i|)\right](\bm{Q}_i)_{kj}\right\|\\
        &\quad +\left\|\mathrm{concat}\left[\rho(|\tilde{\lambda}'_j-\tilde{\lambda}'_i|)\left(\bm{v}_j-\sum_{k:\tilde{\lambda}'_k\in\mathcal{J}(\tilde{\lambda}'_i)}\bm{Pv}'_k(\bm{Q}_i)_{kj}\right)\right]_{\tilde{\lambda}'_j\in\mathcal{J}(\tilde{\lambda}'_i)}\right\|_\text{F}.\label{eq:temporary}
    \end{align}
    Now we analyze the two terms in \eqref{eq:temporary}. For the first term, 
    \begin{align}
        \notag &\quad\left\|\sum_{k:\tilde{\lambda}'_k\in\mathcal{J}(\tilde{\lambda}'_i)}\bm{Pv}'_k\left[\rho(|\tilde{\lambda}'_j-\tilde{\lambda}'_i|)-\rho(|\tilde{\lambda}'_k-\tilde{\lambda}'_i|)\right](\bm{Q}_i)_{kj}\right\|\\
        &=\left\|\mathrm{concat}\left\{\bm{Pv}'_k\left[\rho(|\tilde{\lambda}'_j-\tilde{\lambda}'_i|)-\rho(|\tilde{\lambda}'_k-\tilde{\lambda}'_i|)\right]\right\}_{\tilde{\lambda}'_k\in\mathcal{J}(\tilde{\lambda}'_i)}(\bm{Q}_i)_{\cdot j}\right\|_\text{F}\label{eq:sss1}\\
        &\leqslant \left\|\mathrm{concat}\left\{\bm{Pv}'_k\left[\rho(|\tilde{\lambda}'_j-\tilde{\lambda}'_i|)-\rho(|\tilde{\lambda}'_k-\tilde{\lambda}'_i|)\right]\right\}_{\tilde{\lambda}'_k\in\mathcal{J}(\tilde{\lambda}'_i)}\right\|_\text{F}\|(\bm{Q}_i)_{\cdot j}\|_2\label{eq:sss2}\\
        &=\left\|\mathrm{concat}\left\{\bm{Pv}'_k\left[\rho(|\tilde{\lambda}'_j-\tilde{\lambda}'_i|)-\rho(|\tilde{\lambda}'_k-\tilde{\lambda}'_i|)\right]\right\}_{\tilde{\lambda}'_k\in\mathcal{J}(\tilde{\lambda}'_i)}\right\|_\text{F}\label{eq:sss3}\\
        &\leqslant \sum_{k:\tilde{\lambda}'_k\in\mathcal{J}(\tilde{\lambda}'_i)}\|\bm{Pv}'_k\|\left|\rho(|\tilde{\lambda}'_j-\tilde{\lambda}'_i|)-\rho(|\tilde{\lambda}'_k-\tilde{\lambda}'_i|)\right|\label{eq:sss4}\\
        &=\sum_{k:\tilde{\lambda}'_k\in\mathcal{J}(\tilde{\lambda}'_i)}\left|\rho(|\tilde{\lambda}'_j-\tilde{\lambda}'_i|)-\rho(|\tilde{\lambda}'_k-\tilde{\lambda}'_i|)\right|.\label{eq:sss5}
    \end{align}
    Here, \eqref{eq:sss1} translates the first term of \eqref{eq:temporary} into the form of matrix multiplication. Then \eqref{eq:sss2} makes use of Lemma~\ref{lemma:product}, and \eqref{eq:sss3} further uses the fact that $\bm{Q}_i$ is orthogonal. Finally, \eqref{eq:sss5} stems from the fact that $\bm{v}'_k$ is a normalized eigenvector. Regarding \eqref{eq:sss5}, we may discuss two cases. If both $|\tilde{\lambda}'_j-\tilde{\lambda}'_i|\leqslant \delta$ and $|\tilde{\lambda}'_k-\tilde{\lambda}'_i|\leqslant \delta$, then
    \begin{align}
        \notag&\quad \sum_{k:\tilde{\lambda}'_k\in\mathcal{J}(\tilde{\lambda}'_i)}\left|\rho(|\tilde{\lambda}'_j-\tilde{\lambda}'_i|)-\rho(|\tilde{\lambda}'_k-\tilde{\lambda}'_i|)\right|\\
        &\leqslant J_\rho \sum_{k:\tilde{\lambda}'_k\in\mathcal{J}(\tilde{\lambda}'_i)}\left||\tilde{\lambda}'_j-\tilde{\lambda}'_i|-|\tilde{\lambda}'_k-\tilde{\lambda}'_i|\right|\\
        &\leqslant 2\delta J_\rho |\mathcal{J}(\tilde{\lambda}'_i)|.
    \end{align}
    If at least one of $|\tilde{\lambda}'_j-\tilde{\lambda}'_i|$ and $|\tilde{\lambda}'_k-\tilde{\lambda}'_i|$ exceeds $\delta$, we may assume without loss of generality that $|\tilde{\lambda}'_j-\tilde{\lambda}'_i|>\delta$. Then $\rho(|\tilde{\lambda}'_j-\tilde{\lambda}'_i|)=\rho(\delta)=0$ by continuity of $\rho$, and we still have
    \begin{align}
        \notag&\quad \sum_{k:\tilde{\lambda}'_k\in\mathcal{J}(\tilde{\lambda}'_i)}\left|\rho(|\tilde{\lambda}'_j-\tilde{\lambda}'_i|)-\rho(|\tilde{\lambda}'_k-\tilde{\lambda}'_i|)\right|\\
        &= \sum_{k:\tilde{\lambda}'_k\in\mathcal{J}(\tilde{\lambda}'_i)}\left|\rho(\delta)-\rho(|\tilde{\lambda}'_k-\tilde{\lambda}'_i|)\right|\\
        &\leqslant J_\rho \sum_{k:\tilde{\lambda}'_k\in\mathcal{J}(\tilde{\lambda}'_i)}\left|\delta-|\tilde{\lambda}'_k-\tilde{\lambda}'_i|\right|\\
        &\leqslant 2\delta J_\rho |\mathcal{J}(\tilde{\lambda}'_i)|.
    \end{align}
    Therefore, we conclude that
    \begin{align}
        \left\|\sum_{k:\tilde{\lambda}'_k\in\mathcal{J}(\tilde{\lambda}'_i)}\bm{Pv}'_k\left[\rho(|\tilde{\lambda}'_j-\tilde{\lambda}'_i|)-\rho(|\tilde{\lambda}'_k-\tilde{\lambda}'_i|)\right](\bm{Q}_i)_{kj}\right\|\leqslant 2\delta J_\rho |\mathcal{J}(\tilde{\lambda}'_i)|,
    \end{align}
    or
    \begin{align}
        \sum_{j:\tilde{\lambda}'_j\in\mathcal{J}(\tilde{\lambda}'_i)}\left\|\sum_{k:\tilde{\lambda}'_k\in\mathcal{J}(\tilde{\lambda}'_i)}\bm{Pv}'_k\left[\rho(|\tilde{\lambda}'_j-\tilde{\lambda}'_i|)-\rho(|\tilde{\lambda}'_k-\tilde{\lambda}'_i|)\right](\bm{Q}_i)_{kj}\right\|\leqslant 2\delta J_\rho |\mathcal{J}(\tilde{\lambda}'_i)|^2.\label{eq:intermediate1}
    \end{align}
    For the second term of \eqref{eq:temporary}, we have
    \begin{align}
        \notag&\quad \left\|\mathrm{concat}\left[\rho(|\tilde{\lambda}'_j-\tilde{\lambda}'_i|)\left(\bm{v}_j-\sum_{k:\tilde{\lambda}'_k\in\mathcal{J}(\tilde{\lambda}'_i)}\bm{Pv}'_k(\bm{Q}_i)_{kj}\right)\right]_{\tilde{\lambda}'_j\in\mathcal{J}(\tilde{\lambda}'_i)}\right\|_\text{F}\\
        &\leqslant \left\|\mathrm{concat}\left[\bm{v}_j-\sum_{k:\tilde{\lambda}'_k\in\mathcal{J}(\tilde{\lambda}'_i)}\bm{Pv}'_k(\bm{Q}_i)_{kj}\right]_{\tilde{\lambda}'_j\in\mathcal{J}(\tilde{\lambda}'_i)}\right\|_\text{F}\label{eq:aaaa1}\\
        &=\left\|\mathrm{concat}\left[\bm{v}_j\right]_{\tilde{\lambda}'_j\in\mathcal{J}(\tilde{\lambda}'_i)}-\mathrm{concat}\left[\bm{Pv}'_j\right]_{\tilde{\lambda}'_j\in\mathcal{J}(\tilde{\lambda}'_i)}\bm{Q}_i\right\|_\text{F}.\label{eq:aaaa2}
    \end{align}
    Here, \eqref{eq:aaaa1} uses the fact that $\rho(|\tilde{\lambda}'_j-\tilde{\lambda}'_i|)\in[0,1]$, and \eqref{eq:aaaa2} rewrites \eqref{eq:aaaa1} into matrix multiplication. We further transform \eqref{eq:aaaa2} into
    \begin{align}
        \notag&\quad\left\|\mathrm{concat}\left[\bm{v}_j\right]_{\tilde{\lambda}'_j\in\mathcal{J}(\tilde{\lambda}'_i)}-\mathrm{concat}\left[\bm{Pv}'_j\right]_{\tilde{\lambda}'_j\in\mathcal{J}(\tilde{\lambda}'_i)}\bm{Q}_i\right\|_\text{F}\\
        &\leqslant \left\|\mathrm{concat}\left[\bm{v}_j\right]_{\tilde{\lambda}'_j\in\mathcal{J}(\tilde{\lambda}'_i)}\bm{Q}_i^T-\mathrm{concat}\left[\bm{Pv}'_j\right]_{\tilde{\lambda}'_j\in\mathcal{J}(\tilde{\lambda}'_i)}\right\|_\text{F}\|\bm{Q}_i^T\|_2\label{eq:bbb1}\\
        &=\left\|\mathrm{concat}\left[\bm{Pv}'_j\right]_{\tilde{\lambda}'_j\in\mathcal{J}(\tilde{\lambda}'_i)}-\mathrm{concat}\left[\bm{v}_j\right]_{\tilde{\lambda}'_j\in\mathcal{J}(\tilde{\lambda}'_i)}\bm{Q}_i^T\right\|_\text{F},\label{eq:bbb2}
    \end{align}
    in which \eqref{eq:bbb1} makes use of Lemma~\ref{lemma:product}, and \eqref{eq:bbb2} uses the fact that the spectral norm of an orthogonal matrix is always 1. Now, we may apply Lemma~\ref{lemma:davis-kahan} on \eqref{eq:bbb2} to find that there exists $\bm{Q}_i\in O(|\mathcal{J}(\tilde{\lambda}'_i)|)$, such that
    \begin{align}
        \notag&\quad\left\|\mathrm{concat}\left[\bm{Pv}'_j\right]_{\tilde{\lambda}'_j\in\mathcal{J}(\tilde{\lambda}'_i)}-\mathrm{concat}\left[\bm{v}_j\right]_{\tilde{\lambda}'_j\in\mathcal{J}(\tilde{\lambda}'_i)}\bm{Q}_i^T\right\|_\text{F}\\
        &\leqslant \frac{\sqrt{8}}{\delta}\min\left\{\sqrt{|\mathcal{J}(\tilde{\lambda}'_i)|}\cdot\|\bm{PL}'\bm{P}^T-\bm{L}\|_2, \|\bm{PL}'\bm{P}^T-\bm{L}\|_\text{F}\right\}. \label{eq:temptemp}
    \end{align}
    To arrive at \eqref{eq:temptemp}, we exploit the fact that at boundaries of $\mathcal{J}(\tilde{\lambda}'_i)$ (assumed to be $\tilde{\lambda}'_{J_{\ell-1}+1}$ and $\tilde{\lambda}'_{J_\ell}$), we always have $\tilde{\lambda}'_{J_{\ell-1}+1}-\tilde{\lambda}'_{J_{\ell-1}}>\delta$ and $\tilde{\lambda}'_{J_\ell+1}-\tilde{\lambda}'_{J_\ell}>\delta$. Thus, we end up finding that
    \begin{align}
        \notag&\quad \left\|\mathrm{concat}\left[\rho(|\tilde{\lambda}'_j-\tilde{\lambda}'_i|)\left(\bm{v}_j-\sum_{k:\tilde{\lambda}'_k\in\mathcal{J}(\tilde{\lambda}'_i)}\bm{Pv}'_k(\bm{Q}_i)_{kj}\right)\right]_{\tilde{\lambda}'_j\in\mathcal{J}(\tilde{\lambda}'_i)}\right\|_\text{F}\\
        &\leqslant \frac{\sqrt{8}}{\delta}\min\left\{\sqrt{|\mathcal{J}(\tilde{\lambda}'_i)|}\cdot\|\bm{L}-\bm{PL}'\bm{P}^T\|_2, \|\bm{L}-\bm{PL}'\bm{P}^T\|_\text{F}\right\}.\label{eq:intermediate2}
    \end{align}
    Plugging equations~\eqref{eq:intermediate1} and~\eqref{eq:intermediate2} into~\eqref{eq:temporary}, we find that $\exists \bm{Q}_i\in O(|\mathcal{J}(\tilde{\lambda}'_i)|)$, such that
    \begin{align}
        \notag &\quad \left\|\mathrm{concat}\left[\bm{v}_j\rho(|\tilde{\lambda}'_j-\tilde{\lambda}'_i|)\right]_{\tilde{\lambda}'_j\in\mathcal{J}(\tilde{\lambda}'_i)}-\mathrm{concat}\left[\bm{Pv}'_j\rho(|\tilde{\lambda}'_j-\tilde{\lambda}'_i|)\right]_{\tilde{\lambda}'_j\in\mathcal{J}(\tilde{\lambda}'_i)}\bm{Q}_i\right\|_\text{F}\\
        &\leqslant 2\delta J_\rho |\mathcal{J}(\tilde{\lambda}'_i)|^2+\frac{\sqrt{8}}{\delta}\min\left\{\sqrt{|\mathcal{J}(\tilde{\lambda}'_i)|}\cdot\|\bm{L}-\bm{PL}'\bm{P}^T\|_2, \|\bm{L}-\bm{PL}'\bm{P}^T\|_\text{F}\right\}\\
        &\leqslant 2n^2\delta J_\rho+\frac{\sqrt{8}}{\delta}\|\bm{L}-\bm{PL}'\bm{P}^T\|_\text{F}.
    \end{align}
    Therefore,
    \begin{align}
        \notag&\quad\min_{\bm{Q}_i\in O(n)}\left\|\mathrm{concat}\left[\bm{v}_j\rho(|\tilde{\lambda}'_j-\tilde{\lambda}'_i|)\right]_{j=1}^n-\mathrm{concat}\left[\bm{Pv}'_j\rho(|\tilde{\lambda}'_j-\tilde{\lambda}'_i|)\right]_{j=1}^n\bm{Q}_i\right\|_\text{F}\\
        &\leqslant 2n^2\delta J_\rho+\frac{\sqrt{8}}{\delta}\|\bm{L}-\bm{PL}'\bm{P}^T\|_\text{F}.\label{eq:______}
    \end{align}
    Plugging~\eqref{eq:___third___step} and~\eqref{eq:______} into~\eqref{eq:__fourth__step}, we get
    \begin{align}
        \|f(\tilde{\bm{V}}_i^\text{smooth})-f(\bm{P}\tilde{\bm{V}}_i'^{\text{smooth}})\|\leqslant J_f\left(2nJ_\rho \|\bm{L}-\bm{PL}'\bm{P}^T\|_2+2n^2\delta J_\rho+\frac{\sqrt{8}}{\delta}\|\bm{L}-\bm{PL}'\bm{P}^T\|_\text{F}\right).
    \end{align}
    Combining everything together, we eventually arrive at
    \begin{align}
        \notag\|Z(G)-\bm{P}Z(G')\|_\text{F} &\leqslant nJ_\psi J_\phi\Bigg[(\sqrt{n}+2nJ_\rho J_f)\|\bm{L}-\bm{PL}'\bm{P}^T\|_2\\
        &\qquad\qquad+J_f\left(2n^2\delta J_\rho+\frac{\sqrt{8}}{\delta}\|\bm{L}-\bm{PL}'\bm{P}^T\|_\text{F}\right)\Bigg].\label{eq:semi-final}
    \end{align}
    By choosing a $\delta$ value that minimizes the RHS of equation~\eqref{eq:semi-final}, we get
    \begin{align}
        \notag\|Z(G)-\bm{P}Z(G')\|_\text{F} &\leqslant nJ_\psi J_\phi\big[(\sqrt{n}+2nJ_\rho J_f)\|\bm{L}-\bm{PL}'\bm{P}^T\|_2\\
        &\qquad\qquad+4\sqrt[4]{2}J_f\sqrt{J_\rho}n\|\bm{L}-\bm{PL}'\bm{P}^T\|_\text{F}^{1/2}\big],
    \end{align}
    which is our desired final result.
\end{proof}


\section{Experimental details}\label{Sec:appendix_experiment}

\subsection{Dataset descriptions}

The statistics of used datasets in the paper (except for DrugOOD) are summarized in Table~\ref{Table_dataset_statistics}. 

\begin{table}[ht]
\caption{Overview of the datasets used in the paper.}
\label{Table_dataset_statistics}
\begin{center}
\begin{tabular}{ccccccc}
\toprule
\multirow{2}{*}{Dataset} & \multirow{2}{*}{\#Graphs} & Avg. \# & Avg. \# & Prediction & Prediction & \multirow{2}{*}{Metric} \\
 & & nodes & edges & level & task & \\
\midrule
QM9 & 130,000 & 18.0 & 37.3 & graph & regression & Mean Abs. Error\\
ZINC & 12,000 & 23.2 & 24.9 & graph  & regression & Mean Abs. Error\\
Alchemy & 202,579 & 10.0 & 10.4 & graph & regression & Mean Abs. Error\\
PCQM-Contact & 529,434 & 30.1 & 61.0 & inductive link & link ranking & MRR\\
CLUSTER & 12,000 & 117.20 & 4,301.72 & node & classification & Accuracy\\
PATTERN & 14,000 & 117.47 & 4,749.15 & node & classification & Accuracy\\
ogbg-molhiv & 41,127 & 25.5 & 27.5 & graph  & classification & AUROC \\
\bottomrule
\end{tabular}
\end{center}
\end{table}

\subsection{Implementation details}

\subsubsection{Architecture design}

To implement OGE-Aug practically, the central issue is to choose a proper orthogonal-group invariant encoder $f$ in equation \eqref{eq:smooth1}. In our experiments, we uniformly adopt a point cloud network architecture similar to the one proposed in \citep{finkelshtein2022simple}. We provide the detailed implementation in Algorithm~\ref{alg:oge-aug}. Here, $\mathrm{Linear}^{\bm{Q},\bm{b}}_{\text{shape}_1\rightarrow\text{shape}_2}$ or $\mathrm{Linear}^{\bm{Q}}_{\text{shape}_1\rightarrow\text{shape}_2}$ means a linear transformation operating on the last dimension of $\text{shape}_1$ and transforming it into $\text{shape}_2$, either with or without bias $\bm{b}$. In PyTorch, such operations would translate to \texttt{nn.Linear} modules. The operator $\mathrm{matmul}$ operates similarly to \texttt{torch.matmul}. The function $\rho(x)$ takes the form
\begin{align}
    \rho(x)=\left\{\begin{array}{ll}
    \frac{1}{2}\left(1+\cos \frac{\pi x}{\delta}\right), & 0\leqslant x\leqslant \delta,\\
    0, & x>\delta,
    \end{array}\right.\label{eq:rho_shape_actual}
\end{align}
where $\delta$ is a hyperparameter.

\SetKwComment{Comment}{\# }{}

\begin{algorithm2e}[htbp]
\caption{Practical implementation of OGE-Aug.}\label{alg:oge-aug}
\KwData{Node features $\bm{X}\in \mathbb{R}^{n\times d}$, the matrix of Laplacian eigenvectors $\bm{V}=(\bm{v}_1,\ldots, \bm{v}_n)\in$ $\mathbb{R}^{n\times n}$, and $\tilde{\bm{V}}_1^\text{smooth},\ldots, \tilde{\bm{V}}_n^\text{smooth}\in\mathbb{R}^{n\times n}$ as defined in equation \eqref{eq:ineedit}.}
\KwResult{Node feature augmentations $\bm{Z}\in\mathbb{R}^{n\times h}$.}

\SetKwBlock{Funccc}{\textbf{(a) Preparation.} \textnormal{Given weight matrices $\bm{Q}_0^\mathrm{init}\in\mathbb{R}^{d\times h},\bm{b}_0^\mathrm{init}\in\mathbb{R}^h, \bm{Q}_1^\mathrm{init}, \bm{Q}_2^\mathrm{init}\in\mathbb{R}^{1\times h}$,}}{}
\SetAlgoLined
\Funccc{
$\mathbf{W}^{(0)}\leftarrow\mathrm{Linear}^{\bm{Q}^\mathrm{init}_0,\bm{b}^\mathrm{init}_0}_{(\cdot,d)\rightarrow (\cdot, h)}(\bm{X})$ \Comment*[r]{$\mathbf{W}^{(0)}\in\mathbb{R}^{n\times h}\quad\quad$}
$\mathbf{W}^{(1)}\leftarrow\mathrm{Linear}^{\bm{Q}^\mathrm{init}_1}_{(\cdot, \cdot, 1)\rightarrow(\cdot, \cdot, h)}(\bm{V}.\text{unsqueeze}(-1))$ \Comment*[r]{$\mathbf{W}^{(1)}\in\mathbb{R}^{n\times n\times h}\quad$}
$\mathbf{W}^{(2)}_{a,j,k,:}\leftarrow\mathrm{Linear}^{\bm{Q}^\mathrm{init}_2}_{1\rightarrow h}\left[(\tilde{\bm{V}}_j^\text{smooth})_{ak}(\tilde{\bm{V}}_k^\text{smooth})_{aj}\right]$ \Comment*[r]{$\mathbf{W}^{(2)}\in\mathbb{R}^{n\times n\times n\times h}$}
}

\SetKwBlock{Funcc}{\textbf{(b) Updates.} \textnormal{Alternately apply the following two types of layers for $N$ times.}}{}
\SetAlgoLined

\Funcc{\SetKwBlock{Func}{\textbf{(i) Tensor product layer.} \textnormal{Given input $\mathbf{W}^{(0)},\mathbf{W}^{(1)},\mathbf{W}^{(2)}$, weight matrices $\bm{Q}^\mathrm{prod}_0,\bm{Q}^\mathrm{prod}_1,$ $\bm{Q}^\mathrm{prod}_2,\bm{R}^\mathrm{prod}_0,\bm{R}^\mathrm{prod}_1,\bm{R}^\mathrm{prod}_2\in\mathbb{R}^{h\times h},\bm{b}^\mathrm{prod}_0\in\mathbb{R}^{ h}$ and $\bm{c}\in\mathbb{R}^{3\times 3}$, }}{}
\SetAlgoLined
\Func{
  \textcircled{1} $\mathbf{W}^{(1)}_\mathrm{norm}\leftarrow\mathrm{Normalize}(\mathbf{W}^{(1)},\text{dim}=1)$\;
  \textcircled{2} $\mathbf{W}^{(2)}_\mathrm{norm}\leftarrow\mathrm{Normalize}(\mathbf{W}^{(2)},\text{dim}=(1,2))$\;
  \textcircled{3} $\tilde{\mathbf{W}}^{(0)},\tilde{\mathbf{W}}^{(1)},\tilde{\mathbf{W}}^{(2)}\leftarrow \sigma\left(\mathrm{Linear}^{\bm{Q}_0^\mathrm{prod},\bm{b}_0^\mathrm{prod}}_{(\cdot, h)\rightarrow(\cdot, h)}(\mathbf{W}^{(0)})\right),\mathrm{Linear}^{\bm{Q}_1^\mathrm{prod}}_{(\cdot, \cdot, h)\rightarrow(\cdot, \cdot, h)}(\mathbf{W}^{(1)}_\mathrm{norm}),$ $\mathrm{Linear}^{\bm{Q}_2^\mathrm{prod}}_{(\cdot,\cdot, \cdot, h)\rightarrow(\cdot,\cdot, \cdot, h)}(\mathbf{W}^{(2)}_\mathrm{norm})$, where $\sigma(\cdot)$ is a normalization layer followed by element-wise SiLU\;
  \textcircled{4} $\mathbf{W}^{(0)}_{ij}\leftarrow\mathbf{W}_{ij}^{(0)}+\mathrm{matmul}\Big[c_{00}\mathbf{W}^{(0)}_{ij}\tilde{\mathbf{W}}^{(0)}_{ij}+c_{01}\sum_k\mathbf{W}^{(1)}_{ikj}\tilde{\mathbf{W}}^{(1)}_{ikj}+c_{02}\sum_{k,\ell}\mathbf{W}^{(2)}_{ik\ell j}\tilde{\mathbf{W}}^{(2)}_{ik\ell j},\bm{R}_0^\mathrm{prod}\Big]$\;
  \textcircled{5} $\mathbf{W}^{(1)}_{ikj}\leftarrow\mathbf{W}^{(1)}_{ikj}+\mathrm{matmul}\Big[c_{10}\mathbf{W}^{(1)}_{ikj}\tilde{\mathbf{W}}^{(0)}_{ij}+c_{12}\sum_\ell \mathbf{W}^{(1)}_{i\ell j}\tilde{\mathbf{W}}^{(2)}_{ik\ell  j},\bm{R}^{\mathrm{prod}}_1\Big]$\;
  \textcircled{6} $\mathbf{W}^{(2)}_{ik\ell j}\leftarrow\mathbf{W}^{(2)}_{ik\ell j}+\mathrm{matmul}\Big[c_{20}\mathbf{W}^{(2)}_{ik\ell j}\tilde{\mathbf{W}}^{(0)}_{ij}+c_{11}\rho^2(|\tilde{\lambda}_k-\tilde{\lambda}_\ell|)\mathbf{W}^{(1)}_{ikj}\tilde{\mathbf{W}}^{(1)}_{i\ell j}+c_{22}\rho^2(|\tilde{\lambda}_k-\tilde{\lambda}_\ell|)\sum_m\mathbf{W}^{(2)}_{ikmj}\tilde{\mathbf{W}}^{(2)}_{im\ell j}, \bm{R}^\mathrm{prod}_2\Big]$\;
}

\SetKwBlock{Func}{\textbf{(ii) Message passing layer.} \textnormal{Given input $\mathbf{W}^{(0)},\mathbf{W}^{(1)},\mathbf{W}^{(2)}$, adjacency matrix $\bm{A}$ and weight matrices $\bm{Q}^\mathrm{msg}_0,\bm{Q}^\mathrm{msg}_1,\bm{Q}^\mathrm{msg}_2\in\mathbb{R}^{h\times h},\bm{b}^\mathrm{msg}_0\in\mathbb{R}^{ h}$, }}{}
\SetAlgoLined
\Func{
  \textcircled{1} $\mathbf{W}^{(1)}_\mathrm{norm}\leftarrow\mathrm{Normalize}(\mathbf{W}^{(1)},\text{dim}=1)$\;
  \textcircled{2} $\mathbf{W}^{(2)}_\mathrm{norm}\leftarrow\mathrm{Normalize}(\mathbf{W}^{(2)},\text{dim}=(1,2))$\;
  \textcircled{3} $\tilde{\mathbf{W}}^{(0)},\tilde{\mathbf{W}}^{(1)},\tilde{\mathbf{W}}^{(2)}\leftarrow \sigma\left(\mathrm{Linear}^{\bm{Q}_0^\mathrm{msg},\bm{b}_0^\mathrm{msg}}_{(\cdot, h)\rightarrow(\cdot, h)}(\mathbf{W}^{(0)})\right),\mathrm{Linear}^{\bm{Q}_1^\mathrm{msg}}_{(\cdot, \cdot, h)\rightarrow(\cdot, \cdot, h)}(\mathbf{W}^{(1)}_\mathrm{norm}),$ $\mathrm{Linear}^{\bm{Q}_2^\mathrm{msg}}_{(\cdot,\cdot, \cdot, h)\rightarrow(\cdot,\cdot, \cdot, h)}(\mathbf{W}^{(2)}_\mathrm{norm})$, where $\sigma(\cdot)$ is a normalization layer followed by element-wise SiLU\;
  \textcircled{4} $\mathbf{W}^{(0)}_{i:}\leftarrow\mathbf{W}^{(0)}_{i:}+\sum_k A_{ik}\tilde{\mathbf{W}}^{(0)}_{k:}$\;
  \textcircled{5} $\mathbf{W}^{(1)}_{i::}\leftarrow\mathbf{W}^{(1)}_{i::}+\sum_k A_{ik}\tilde{\mathbf{W}}^{(1)}_{k::}$\;
  \textcircled{6} $\mathbf{W}^{(2)}_{i:::}\leftarrow\mathbf{W}^{(2)}_{i:::}+\sum_k A_{ik}\tilde{\mathbf{W}}^{(2)}_{k:::}$\;
}

}
\SetKwBlock{Funccc}{\textbf{(c) Output.} \textnormal{$\bm{Z}\leftarrow\mathbf{W}^{(0)}$.}}{}
\SetAlgoLined
\Funccc{}

\end{algorithm2e}

We now discuss the complexity of Algorithm~\ref{alg:oge-aug} as well as its connections to our theoretically proposed OGE-Aug (Definition~\ref{def:oge-aug}). It is not hard to notice that the most computationally costly steps of Algorithm~\ref{alg:oge-aug} are those to compute $\rho^2(|\tilde{\lambda}_k-\tilde{\lambda}_\ell|)\sum_m\mathbf{W}^{(2)}_{ikmj}\tilde{\mathbf{W}}^{(2)}_{im\ell j}$ and $\sum_k A_{ik}\tilde{\mathbf{W}}^{(2)}_{k:::}$. If we use dense matrices to store all the necessary data, the time complexity to compute those two terms are $O(n^4)$ and $O(n^2m)$, where $n$ and $m$ refer to the number of nodes and edges of $G$, respectively. Nevertheless, since the smoothing function $\rho(\cdot)$ is only non-zero when its argument is sufficiently close to zero, we find that $\mathbf{W}^{(2)}_{i::j}$ is a sparse matrix with only $O(n\max_j \mu_j)$ non-zero elements, for each $i=1,\ldots, n$ and $j=1,\ldots, h$. Here, $\max_j\mu_j$ means the maximum multiplicity of $G$'s Laplacian eigenvalues. Therefore, by storing $\mathbf{W}^{(2)}$ as a sparse matrix, the above two terms can be computed in $O(n^2\max_j\mu^2_j)$ and $O(m\max_j\mu_j)$ time respectively, resulting a practical time complexity of $O((n^2\max_j\mu_j+m)\cdot \max_j\mu_j)$, which is generally lower than $O(n^3)$. 

We remark that although Algorithm~\ref{alg:oge-aug} uses only tensors up to second order, it is not hard to generalize Algorithm~\ref{alg:oge-aug} to accommodate higher-order tensors based on $\tilde{\bm{V}}_1^\text{smooth},\ldots,\tilde{\bm{V}}_n^\text{smooth}$, resulting in a model with higher complexities and better expressive power. When the tensor order reaches $n$, our implementation of OGE-Aug can produce universally expressive graph representations, recovering our theoretical result. Since this would entail an unaffordable complexity of $O(n\cdot n^n)=n\exp(\tilde{O}(n))$, Algorithm~\ref{alg:oge-aug} is adopted practically instead, at the cost of some expressivity.

Finally, we point out that Algorithm~\ref{alg:oge-aug} does not tightly follow equation~\eqref{eq:smooth1}, in that (i) apart from using $\bm{V}_1^\text{smooth},\ldots,\bm{V}_K^\text{smooth}$ (to build second-order tensors), Algorithm~\ref{alg:oge-aug} also uses information directly from the raw Laplacian eigenvectors (to build first-order tensors), and that (ii) Algorithm~\ref{alg:oge-aug} allows mixing of $\bm{V}^\text{smooth}_j$ with different $j$. Despite those differences, Algorithm~\ref{alg:oge-aug} maintains the key idea of OGE-Aug: only information from two Laplacian eigenspaces whose corresponding eigenvalues are ``not too far away'' from each other would be multiplied into $\mathbf{W}^{(2)}$, and the algorithm has no explicit dependence on the multiplicities of Laplacian eigenvalues. Thus, the stability result demonstrated in Proposition~\ref{prop:stability_bound} can similarly hold for Algorithm~\ref{alg:oge-aug}, though the accurate bound may be different.

\subsubsection{Other details of the practical implementation}
We implement OGE-Aug with the PyGHO library~\citep{wang2023pytorch}. To integrate OGE-Aug with other base models including MPNN and graph transformers, we also implement our methods building on the GraphGPS~\citep{rampavsek2022recipe} code base, where we build OGE-Aug as a plug-and-play module. The module takes in Laplacians as inputs and processes the eigenvalues/eigenvectors using \textit{\# PE layers} with dimension \textit{PE hidden dim}, and outputs an embedding with dimension \textit{PE dim}; see \Cref{Table_experiment_settings} for detailed settings. In this module, we use permutation-equivariant set function~\citep{zaheer2017deep} to process the eigenvalues and multiply the eigenvalue embeddings to the eigenvectors. Moreover, we also multiply eigenvectors with eigenvectors to initialize the higher order representations. After that, this module will product each node's representation with its neighbors' and update the representation iteratively. The embedding is then combined with other node features and other optional positional encodings, then fed jointly into downstream layers (which consist of various GNN and graph transformer modules). Therefore, OGE-Aug can be either used solely or integrated easily with arbitrary backbones.

We also implement a version where OGE-Aug modules act on the embeddings of nodes and edges, which can be viewed as operating on weighted or latent Laplacians incorporating node and edge features. However, we experimentally find that processing the original Laplacians with OGE-Aug and encoding the node/edge features separately via other encoders (as explained above) yields better performance.  

In addition, to make OGE-Aug more robust, we add a small-scale noise (typically a Gaussian noise with mean zero and variance ${10}^{-5}$) to the Laplacians in the training process. We also randomly permute the Laplacians and do inverse permutation to the output eigenvectors to simulate the noise caused by the permutation and the numerical algorithm. We use the original Laplacians in the inference stage.





\begin{table}[ht]
\caption{Hyperparameters of the experiments.}
\label{Table_experiment_settings}
\begin{center}
\begin{tabular}{lcccc}
\toprule
Hyperparameters & QM9 & ZINC & Alchemy & PCQM-Contact\\
\midrule
\# Layers & 10 & 10 & 16 & 6 \\
Hidden dim & 64 & 64 & 128 & 96 \\
MPNN & GINE & GINE & GINE & GatedGCN \\
Attention & Transformer$^\dagger$ & Transformer$^*$ & - & Transformer \\
\# Heads & 4 & 4 & - & 4 \\
Dropout & 0 & 0 & 0 & 0 \\
Attention dropout & 0.2 & 0.5 & - & 0.1 \\
Graph pooling & sum & sum & sum & edge dot \\
\midrule
Positional encoding & OGE-Aug(29) & OGE-Aug(37) & OGE-Aug(12) & OGE-Aug + LapPE \\
PE hidden dim & 64 & 64 & 64 & 32 \\
PE dim & 28 & 28 & 28 & 16 \\
PE \# layer & 4 & 4 & 4 & 3 \\
\midrule
Batch size & 256 & 32 & 128 & 64 \\
Learning rate & 0.001 & 0.001 & 0.001 & 0.0005 \\
\# Epochs & 500 & 2000 & 1000 & 100 \\
\# Warmup epochs & 50 & 50 & 50 & 10 \\
Weight decay & 1e-5 & 1e-5 & 1e-5 & 0 \\
\midrule
\# Parameters & 783,249 & 617,677 & 1,968,352 & 845,632 \\
Time (epoch/total) & 139s/19.3h & 28s/15.6h & 5s/1.4h & 1541s/42.8h \\
\bottomrule
\end{tabular}
\end{center}
\end{table}

\subsection{Experimental settings}

As explained earlier, we integrate our OGE-Aug with the GraphGPS code base, and thus also follow their experimental settings. With only mild hyperparameter search, we achieve SOTA or highly competitive results on all datasets. The adopted hyperparameters in our experiments presented in the main text are summarized in \Cref{Table_experiment_settings} and \Cref{Table_experiment_settings_2}.

The $^\dagger$ for QM9 suggests that experiments on these four targets $U_0, U, G, H$ are conducted using the PyGHO code version without GraphGPS. $^*$ for ZINC means that the transformer is not necessary - actually we can achieve highly competitive results even without global attention. When we use transformers, we reduce the PE hidden dimension to 32, PE dimension to 16, and PE \# layers to 3, resulting 505905 total number of parameters and 14.9h total training time, which are both less than the case without transformers.

\begin{table}[ht]
\caption{Hyperparameters of the experiments (continued).}
\label{Table_experiment_settings_2}
\begin{center}
\begin{tabular}{lcccc}
\toprule
Hyperparameters & CLUSTER & PATTERN & ogbg-molhiv & DrugOOD\\
\midrule
\# Layers & 16 & 7 & 10 & 6 \\
Hidden dim & 64 & 64 & 64 & 96 \\
MPNN & GatedGCN & GatedGCN & GatedGCN & GINE \\
Attention & Transformer & Transformer & Transformer & - \\
\# Heads & 8 & 4 & 4 & - \\
Dropout & 0.1 & 0 & 0.05 & 0.1 \\
Attention dropout & 0.5 & 0.5 & 0.5 & - \\
Graph pooling & - & - & mean & mean \\
\midrule
Positional encoding & OGE-Aug + ESLapPE & OGE-Aug + LapPE & OGE-Aug + RWSE & OGE-Aug(full) \\
PE hidden dim & 24 & 20 & 16 & 4 \\
PE dim & 32 & 20 & 16 & 8 \\
PE \# layer & 3 & 3 & 2 & 1 \\
\midrule
Batch size & 16 & 32 & 16 & 4 \\
Learning rate & 0.0005 & 0.0005 & 0.0001 & 0.0001 \\
\# Epochs & 100 & 100 & 100 & 200 \\
\# Warmup epochs & 5 & 5 & 5 & 5 \\
Weight decay & 1e-5 & 1e-5 & 1e-5 & 1e-4 \\
\midrule
\# Parameters & 934,798 & 424,517 & 574,673 & 183,533\\
Time (epoch/total) & 226s/6.3h & 178s/4.9h & 387s/10.8h & 725s/40.3h \\
\bottomrule
\end{tabular}
\end{center}
\end{table}

\end{document}